\documentclass{bmvc2k}

\usepackage[ruled,vlined,linesnumbered]{algorithm2e}

\SetCommentSty{mycommfont}
\usepackage{amssymb} 
\usepackage{multicol}
\usepackage{multirow}
\newcommand{\minisection}[1]{\vspace{0.04in} \noindent {\bf #1}\ \ }
\usepackage[final]{pdfpages}

\newcommand{\tabincell}[2]{\begin{tabular}{@{}#1@{}}#2\end{tabular}}

\title{HCV: Hierarchy-Consistency Verification for \\ Incremental Implicitly-Refined Classification}
\addauthor{Kai Wang}{kwang@cvc.uab.es}{1}
\addauthor{Xialei Liu (*\textit{corresponding author})}{xialei@nankai.edu.cn}{2}
\addauthor{Luis Herranz}{lherranz@cvc.uab.es}{1}
\addauthor{Joost van de Weijer}{joost@cvc.uab.es}{1}
\addinstitution{
 Computer Vision Center\\
 Universitat Autònoma de Barcelona\\
 Barcelona, Spain
}
\addinstitution{
 College of Computer Science\\
 Nankai University\\
 Tianjin, China
}

\runninghead{Kai Wang et al.}{HCV for IIRC}
\begin{document}

\maketitle

\begin{abstract}
Human beings learn and accumulate hierarchical knowledge over their lifetime. This knowledge is associated with previous concepts for consolidation and hierarchical construction. However, current incremental learning methods lack the ability to build a concept hierarchy by associating new concepts to old ones. A more realistic setting tackling this problem is referred to as Incremental Implicitly-Refined Classification (IIRC), which simulates the recognition process from coarse-grained categories to fine-grained categories. To overcome forgetting in this benchmark, we propose Hierarchy-Consistency Verification (HCV) as an enhancement to existing continual learning methods. Our method incrementally discovers the hierarchical relations between classes. We then show how this knowledge can be exploited during both training and inference. Experiments on three setups of varying difficulty demonstrate that our HCV module improves performance of existing continual learning methods under this IIRC setting by a large margin. Code is available in \url{https://github.com/wangkai930418/HCV_IIRC}.

\end{abstract}

%-------------------------------------------------------------------------
\section{Introduction}
\label{sec:intro}

In the lifetime of a human being, knowledge is continuously learned and accumulated.  However, deep learning models suffer from knowledge forgetting, also known as catastrophic forgetting~\cite{kirkpatrick2017overcoming,mccloskey1989catastrophic}, when presented with a sequence of tasks. Incremental learning~\cite{parisi2019continual,delange2021continual,masana2020class}, also referred to as continual learning, has been a crucial research direction in computer vision that aims to prevent this forgetting of previous knowledge in neural networks. 

\begin{figure*}[tb]
\begin{center}
\includegraphics[width=\textwidth]{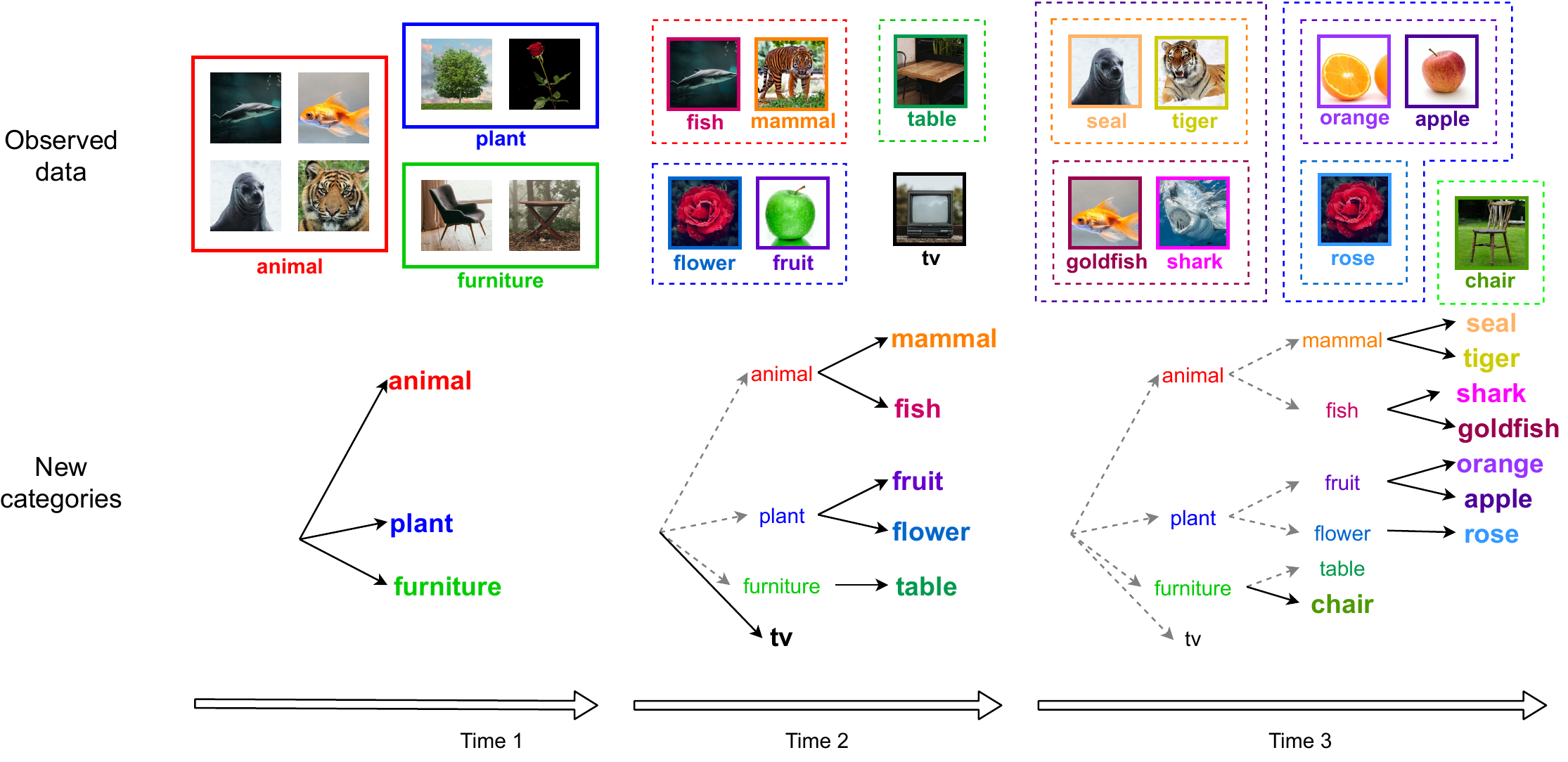}
\end{center}
\vspace{-2mm}
   \caption{Illustration of 3-layer hierarchy IIRC setting. New categories in each training time are annotated by solid pointers, and the hierarchical relationships among old categories and new categories are denoted with dashed arrows. 
 % \vspace{-5mm}
   }
  \vspace{-2mm}
\label{fig:IIRC_setup}
\end{figure*}

Another aspect of human learning is the association of new concepts to old concepts, people construct a hierarchy of knowledge to better consolidate this information. Recently, the IIRC (Incremental Implicitly-Refined Classification) setup~\cite{abdelsalam2020iirc} has been proposed as a novel extended benchmark to evaluate lifelong learning methods in a realistic setting where the construction of hierarchical knowledge is key. On the IIRC benchmark (see Fig.~\ref{fig:IIRC_setup}), each class has multiple granularity levels. But only one label is present at any time, which requires the model to infer whether the related labels have been observed in previous tasks. This setting is much closer to real-life learning, where a learner gradually improves its knowledge of objects (first it labels roses as a plant, later as a flower, and finally a rose). 

Based on this benchmark, Abdelsalam et al.~\cite{abdelsalam2020iirc} adapted and evaluated several state-of-the-art incremental learning methods to address this problem, including iCaRL~\cite{rebuffi2017icarl}, LUCIR~\cite{hou2019learning}, and AGEM~\cite{chaudhry2018efficient}. However, their work does not propose an effective solution specifically designed for the IIRC problem. They do not aim to incrementally learn the hierarchical knowledge that is important to correctly label the data in this setting.  Furthermore, there are also some other limitations in the current version of the IIRC benchmark: (i) The granularity is limited to two layers, while in reality there are often more layers involved (see WordNet~\cite{miller1995wordnet} hierarchy of ImageNet~\cite{imagenet_cvpr09}). 

(ii) The first task always contains a large number of superclasses, which means 
that the learner encounters data from most classes already in these early stages\footnote{The actual setup considers 10 superclasses in the first task, meaning that around 50 (of the total 100) subclasses are seen implicitly during the first task.}. This makes training relatively easy, and the proposed setup less applicable.

To overcome catastrophic forgetting under the IIRC setup, we propose a module called Hierarchy-Consistency Verification (HCV). We aim to explicitly learn in an incremental manner the hierarchical knowledge that underlies the data. While learning new tasks with new super and subclasses, we automatically discover relations, e.g. the class ‘flower’ is a subclass of ‘plant’. Next, we show how this knowledge can be exploited to enhance incremental learning. Principally, in the described example, we would not use images from ‘flower’ as negative examples for the class ‘plant’ (a problem from which the methods in ~\cite{abdelsalam2020iirc} suffer). Next, we show how the hierarchical knowledge can be used at inference time to improve the predictions. Based on these observations, our main contributions are: 

\begin{itemize}
    \item We propose a Hierarchy-Consistency Verification (HCV) module as a solution to the IIRC setup. It incrementally discovers the hierarchical knowledge underlying the data, and exploits this during both training and inference.
    \item We extend the IIRC benchmark to a challenging 3-layer hierarchy on the IIRC-CIFAR dataset. In addition, we propose a much harder setup where the superclasses are distributed uniformly over incremental tasks to test the robustness of different methods.
    \item Experiments show that we successfully acquire hierarchical knowledge, and that exploiting this knowledge leads to significantly improvements of existing incremental learning methods under the IIRC setup (with absolute accuracy gains of 3-20\%). 
\end{itemize}

\section{Related work}
\subsection{Incremental learning}

Incremental learning methods can be categorized into three types~\cite{delange2021continual,masana2020class} as follows.

\minisection{Regularization-based methods.} The first group of techniques add a regularization term to the loss function which impedes changes to the parameters deemed relevant to previous tasks. The difference depends on how to compute the estimation. These methods can be further divided into data-focused~\cite{li2017learning,rannen2017encoder,zhang2020class,jung2016less} and prior-focused~\cite{kirkpatrick2017overcoming,zenke2017continual,liu2018rotate,aljundi2018memory,chaudhry2018riemannian,lee2017overcoming}. Data-focused methods use knowledge distillation from previously learned models. Prior-focused methods estimate the importance of model parameters as a prior for the new model.

\minisection{Parameter isolation methods.} This family focuses on allocating different model parameters to each task. These models begin with a simplified architecture and updated incrementally with new neurons or network layers in order to allocate additional capacity for new tasks. In Piggyback/PackNet~\cite{mallya2018piggyback,mallya2018packnet}, the model learns a separate mask on the weights for each task, whereas in HAT~\cite{serra2018overcoming} masks are applied to the activations. This method is further developed to the case where no forgetting is allowed in~\cite{masana2020ternary}. In general, this branch is restricted to the task-aware (task incremental) setting. Thus, they are more suitable for learning a long sequence of tasks when a task oracle is present. 

\minisection{Replay methods.} This type of methods prevent forgetting by including data from previous tasks, stored either in an episodic memory or via a generative model. There are two main strategies: exemplar rehearsal~\cite{rebuffi2017icarl,chaudhry2018efficient,hou2019learning,wu2019large,liu2020generative} and pseudo-rehearsal~\cite{shin2017continual,wu2018memory}. The former stores a small amount of training samples (also called exemplars) from previous tasks. The latter use generative models learned from previous data distributions to synthesize data.

\subsection{Hierarchical classification and multi-label classification}

Classification problem is normally considered that the categories are not overlapped with each other. However, the concepts in real life are connected to each other with hierarchical information. For example, in ImageNet~\cite{imagenet_cvpr09}, the categories are hierarchized by WordNet~\cite{miller1995wordnet} knowledge. For hierarchical classification~\cite{silla2011survey}, the system groups things according to an explicit hierarchy, which is important to some applications, such as bioinformatics~\cite{freitas2007tutorial} and COVID-19 identification~\cite{pereira2020covid}. Another related area is multi-label classification~\cite{zhang2013review}, where each image is related to multiple labels. Multi-label classification is a generalization of the single-label categorizing problem. In the multi-label problem there is no constraint on how many of the classes the instance can be assigned to. While under this setup, there is no hierarchical constraints among categories. By comparison, on the IIRC setup~\cite{abdelsalam2020iirc}, the hierarchical information is implicitly defined. The developed model for this problem should be able to learn this hierarchy by itself and predict the multiple labels for each instance.

\section{Methodology}
The original work that presented the IIRC setup~\cite{abdelsalam2020iirc} ignores the hierarchical nature of the classes during incremental learning. Consequently, some samples are incorrectly used as negative samples for their superclass labels, potentially resulting in a drop of performance. Here we propose our method to incrementally learn the hierarchy and directly exploit this information to remove said interference. Moreover, we also show how the estimated hierarchy can be exploited at inference time. Our method is general and can be applied to existing methods for incremental learning that can be trained with a binary cross-entropy loss (in experiments we will show results for iCaRL~\cite{rebuffi2017icarl}, and LUCIR~\cite{hou2019learning}).

\subsection{IIRC setup}

Given a series of tasks, each task $t \in [1,T]$ is composed of data $D_t$ from the current class set $C_t$ which can contain both super- and subclasses. During training of task $t$ the model will receive $(x_t^i, y_t^i) \in D^{train}_t,\;y_t^i \in C_t$
where $y_t^i \in \{{u}_t^{i},{v}_t^{i} \}$
is either the subclass ${u}_t^{i}$ or the superclass ${v}_t^{i}$
label of the $i$-th sample $x_t^i$, only one of which is present in $C_t$. In the proposed setup of~\cite{abdelsalam2020iirc}, always first the superclass is learned and later the subclass (like in Fig.~\ref{fig:IIRC_setup}). We will use lowercase $y$ for a one-hot vector, and capital $Y$ to identify a binary vector possibly with multiple non-zero elements.  It is important to note that even if during training only a single label $y_t^i$ is provided, during testing after task $t$ we consider test data $(x_{t}^i, Y_t^i) \in \cup_{j=1}^{t} {D_{j}^{test}}$ where multi-class ground-truth vector $Y_t^i$ contains the subclass and superclass label of sample $x_t^i$ (if these are in $\cup_{1}^{t} C_{t}$\footnote{Some samples might only have a single label since the subclass label is not yet encountered during training.}), i.e., at test time we are expected to predict all non-zero elements in $Y_t^i$.  

To make the common recognition model applied in this multi-class case, in~\cite{abdelsalam2020iirc} they propose to replace the conventional cross-entropy loss by a binary cross-entropy loss:

\begin{equation}
\label{eq:bceloss}
\mathcal{L}_{BCE} =  - \sum_i [y_{t}^{i} \cdot log({{\hat Y}_{t}^{i}}) + (1-y_{t}^{i}) \cdot log(1-{{\hat Y}_{t}^{i}})]
\end{equation}
where ${{\hat Y}_{t}^{i}}=  \mathcal{F}_{t}(x_t^i) $ is the predicted probability vector of sample $x_t^i$, with $\mathcal{F}_{t}$ the current prediction model. They apply this equation to several incremental learning algorithms. However, it should be noted that samples can be wrongly used as a negative sample for their own superclass, because this loss only considers the provided label $y_{t}^{i}$.

We extend the two-layer hierarchy proposed in the original IIRC setup to three layers to verify the effectiveness of our module in more complex scenarios. In this case, each sample contains a three-layer label annotations $Y_t^i$ as: (subclass ${u}_t^{i}$, superclass ${v}_t^{i}$, rootclass ${w}_t^{i}$).

\subsection{HCV: Hierarchy-Consistency Verification}
In the previous section, we discussed that the original solution results in interference during training. The challenge here is that the model should correctly learn the relationship between sub classes $u_t^i$ and super classes ${v}_t^i$, given only the $y_t^i$ information during training time. Here, we propose our method that address this problem. 

\begin{figure*}[tb]
\begin{center}
\includegraphics[width=\textwidth]{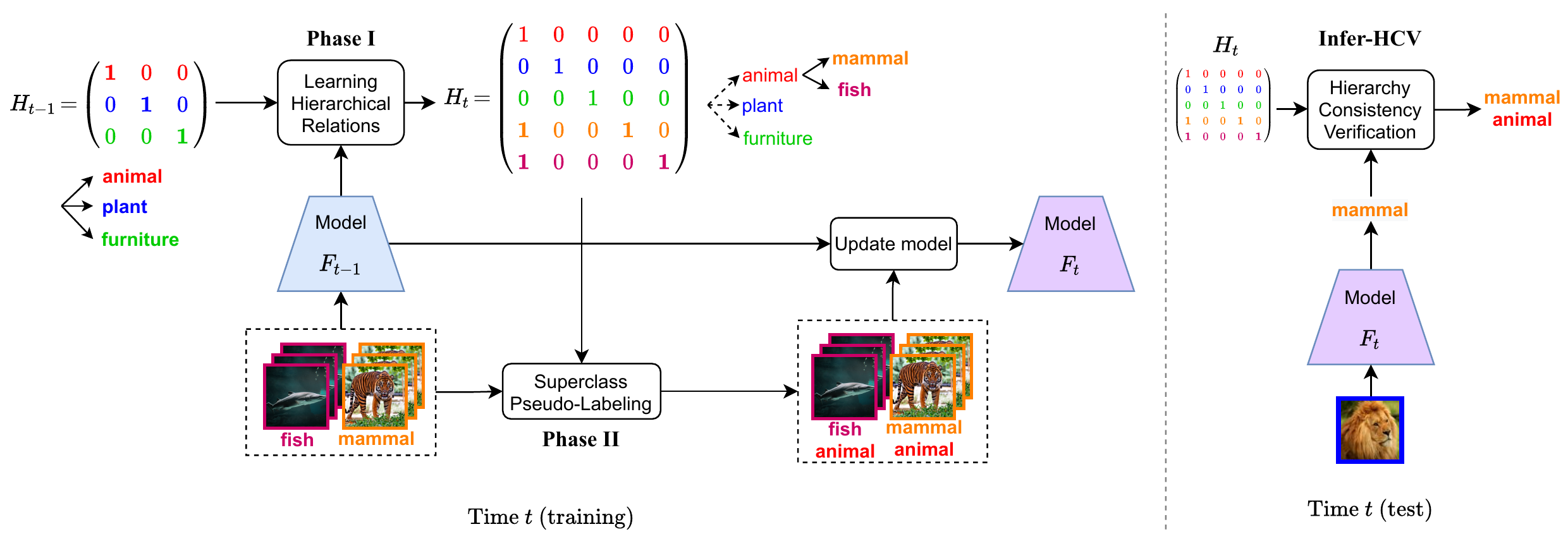}
\end{center}
\vspace{-2mm}
   \caption{Illustration of our method: Hierarchy-Consistency Verification (HCV). At Phase I, hierarchical relations between subclasses and superclasses $H_t$ are acquired using current data. And then at Phase II, the multi-class labels are generated for each instance. Current model is updated with calibrated labels at training time. The hierarchical relations can be applied during inference time as well to further improve the predictions. 
 \vspace{-2mm}
   }
 \vspace{-2mm}
\label{fig:IIRC_HCV}
\end{figure*}

To overcome forgetting under the IIRC setup, we incrementally compute the class hierarchy by estimating the relationship between old and new classes. If a new class is highly related to an old class, we identify it as the subclass of the old class. With this estimated hierarchical knowledge, we verify the hierarchy consistency both during training and inference time to boost the performance of the continual learning models. Our algorithm, called \emph{Hierarchy-Consistency Verification} (HCV), contains two phases which we describe in the following (see also Fig.~\ref{fig:IIRC_HCV}). Moreover, the learned hierarchy is also exploited at inference.

\minisection{Phase I: Learning Hierarchical Relations (LHR).} 

The mission at this stage is to estimate the existing hierarchical relationship between subclasses $u_t^i$ and superclasses ${v}_t^i$. This stage occurs before the training of the current task. 
Supposing we have learned the classifier $\mathcal{F}_{t-1}$ for all previous classes. We could use $\mathcal{F}_{t-1}$ to classify all accessible data $D^{train}_{t}$ for class $y_c$ and produce a prediction vector $p_{y_c}$. 

\begin{equation}
\label{lus}
p_{y_c}= \frac{1}{N} \cdot \sum_{i|y_t^i=y_c} \mathcal{F}_{t-1}(x_t^i)\; \;  \; \; (x_t^i,y_t^i) \in D_t^{train}
\end{equation}
where $N$ is the number of images labeled as $y_c \in C_t$. If the maximum prediction value in $p_{y_c}$ is larger than a threshold $\tau$, we would consider the previous class ${{\bar v}_t^i}$ with the max probability value is the superclass of class $y_t^i$. Based on this prior knowledge learned from previous classifiers, we could construct a hierarchical tree ${H}_t$, which consists of all hierarchical information up to the current task $t$. 

\minisection{Phase II: Superclass Pseudo-Labeling (SPL).} 
After learning the superclasses before training task $t$, we have the hierarchical tree ${H}_t$, which contains all estimated hierarchical information up to the current task. Now we can apply this knowledge at both train and test time.

During training time, if a new class is estimated as a subclass of a specific previous superclass, we assign the estimated superclass label ${{\bar v}_t^i}$ as a \emph{superclass pseudo-label} to the corresponding subclasses label $y_t^i$ (we will use the overline ${{\bar .}}$ to identify that label is estimated). In this way, the estimated multi-class label ${{\bar Y}_{t}^{i}}$ can be represent as:
\begin{equation}
  {{\bar Y}_{t}^{i}} =
    \begin{cases}
      y_t^i & \text{if } y_t^i \text{ has no parents in the hierarchical tree } {H}_t \\
      y_t^i \cup {{\bar v}_t^i} & \text{if } {{\bar v}_t^i} \text{ is the estimated parent of } y_t^i\\
    \end{cases}       
\end{equation}
Then, with the new class label vector ${{\bar Y}_{t}^{i}}$, the binary cross-entropy loss is rewritten as:
\begin{equation}
\label{eq:modified_bceloss}
\mathcal{L}_{BCE} =- \sum_i [{{\bar Y}_{t}^{i}} \cdot log({{\hat Y}_{t}^{i}}) + (1-{{\bar Y}_{t}^{i}}) \cdot log(1-{{\hat Y}_{t}^{i}})]
\end{equation}
For applying our SPL module to continual learning methods, we simply replacing the orginal BCE loss in Eq.~\ref{eq:bceloss} with Eq.~\ref{eq:modified_bceloss}. 

\minisection{Inference with HCV (Infer-HCV).}
At inference time, if a multi-class prediction vector is not consistent with our estimated hierarchical knowledge $H$, we mark it as a wrong prediction (e.g. it estimates a sub and superclass combination that is not in accordance to our hierarchical knowledge captured by $H$). Based on this assumption, we process each prediction ${{\hat Y}_{t}^{i}}$ with $H_t$. If the prediction is in accordance with $H_t$  it remains unchanged. If we need to add labels  to $\hat{Y_{t}^{i}}$ to make it be in accordance to $H_t$ we do so (add subclass or superclass label). If we need to remove labels from ${{\hat Y}_{t}^{i}}$ to reach accordance with $H_t$, we randomly select one of the possible solutions containing the least number of removed labels. See the supplementary material for a visual explanation of Infer-HCV.

\section{Experiments}

\subsection{Experimental setup}
\minisection{Datasets.} We use the same two datasets as in IIRC~\cite{abdelsalam2020iirc}: CIFAR100~\cite{krizhevsky2009learning} and ImageNet~\cite{imagenet_cvpr09}. For CIFAR100, we take the two-level hierarchy split IIRC-CIFAR from IIRC~\cite{abdelsalam2020iirc}, we denote this as IIRC-2-CIFAR. It is composed of 15 superclasses and 100 subclasses. To further explore the performance of incremental learning methods over multi-level hierarchy, we further extend the IIRC-2-CIFAR into a three-level hierarchy dataset IIRC-3-CIFAR with two highest superclasses (we name them as "root"): "animals" and "plants". That accounts 2 rootclasses, 15 superclasses and 100 subclasses. 
For ImageNet, due to its huge amount of data, we collect 100 subclasses according to the hierarchy proposed in IIRC~\cite{abdelsalam2020iirc}. In total there are 10 superclasses and 100 subclasses (including those have no superclass labels). We denote this dataset as IIRC-ImageNet-Subset 
as a simplified version of the original one. The detailed hierarchies and task information are referred to the supplementary material.

\minisection{Incremental task configurations.}For IIRC-2-CIFAR, we adopt the training sequence from IIRC~\cite{abdelsalam2020iirc}, where the first task is with 10 superclasses, in the sequential tasks each with 5 classes. And for IIRC-3-CIFAR, we uniformly distribute the rootclasses and superclasses to form 23 tasks in total, the first task is 7 classes and then the coming tasks are 5 for each. For IIRC-ImageNet-Subset, we have 11 tasks each with 10 classes. Here the superclasses are also uniformly distributed. We want to stress that the uniform distribution of superclasses (and rootclasses) leads to a more challenging setting than proposed in the original IIRC.

\minisection{Baselines and Compared methods.}
We compare the performance of the following variants: 
(1) \textbf{Incremental Joint} learns the model across tasks and the model has access to all the data from previous tasks with complete information (having access to all the label annotations $Y_t$). It serves as the upper bound for comparison.
(2) \textbf{ER-infinite} is similar to \textit{Incremental Joint} but with incomplete information (only access to the current label annotations $y_t$). 
(3) \textbf{iCaRL-CNN} is the original version of incremental learning method iCaRL~\cite{rebuffi2017icarl}. 
(4) \textbf{iCaRL-norm} is the adapted version of iCaRL~\cite{rebuffi2017icarl} with replacement of the distance metric from L2-distance to Cosine similarity. 
(5) \textbf{LUCIR} is the incremental learning method LUCIR~\cite{hou2019learning}. 
(6) \textbf{ER} is the finetuning baseline with 20 image exemplars per class as experience replay. 
(7) \textbf{FT} is the finetuning baseline without image replay.

\minisection{Implementation details.}
For most implementation details, we follow the IIRC configurations~\cite{abdelsalam2020iirc}.
For these three setups, we use the ResNet-32~\cite{he2016deep} as the classification backbone. For model training, we use SGD (momentum=0.9) as optimizer, which is commonly used in continual learning~\cite{mirzadeh2020understanding}. For the IIRC-2-CIFAR and IIRC-3-CIFAR setting, the learning rates begin with 1.0 then decay by 0.1 on the plateau of the  validation performance. For IIRC-ImageNet-Subset, the learning rate starts with 0.5 and decay by 0.1 on the plateau. The number of training epochs is 140, 140 and 100 for IIRC-2-CIFAR, IIRC-3-CIFAR and IIRC-ImageNet-Subset, respectively. For all these three setups, the batch size is 128 and weight decay is 1e-5.

During training, we apply random resized cropping (of size $32\times32$) to both CIFAR100 and ImageNet images. Then a random horizontal flip is applied and followed by a normalization. And for images replay, we keep a fixed number of 20 saved exemplars per class by default.
For evaluation, we adopt the \textit{precision-weighted Jaccard similarity (pw-JS)} proposed in IIRC~\cite{abdelsalam2020iirc}, which integratedly considers both precision and recall indexes. And the threshold $\tau$ is set to 0.6 in all experiments (except in ablation study over it).

\subsection{Experimental results}
\label{sec:expr_results}

\begin{table}[tb]
\begin{center}
\scalebox{0.72}{
\begin{tabular}{|c|c|c|c|c|c|c|c|c|c|}
\hline

Methods & \multicolumn{3}{c|}{iCaRL-CNN} & \multicolumn{3}{c|}{iCaRL-norm} & \multicolumn{3}{c|}{LUCIR} \\
 \hline
  & \multirow{2}{*}{-} & \multirow{2}{*}{\tabincell{c}{+ SPL}} & \multirow{2}{*}{\tabincell{c}{+ SPL \\ + infer HCV}} & \multirow{2}{*}{-} & \multirow{2}{*}{\tabincell{c}{+ SPL}} & \multirow{2}{*}{\tabincell{c}{+ SPL \\ + infer HCV}} & \multirow{2}{*}{-} & \multirow{2}{*}{\tabincell{c}{+ SPL}} & \multirow{2}{*}{\tabincell{c}{+ SPL \\ + infer HCV}} \\
  & &  & &  & &  & & & \\
\hline\hline
{IIRC-2-CIFAR} & 28.4 & 32.7	& \textbf{35.9}	& 24.9	& 29.1& {31.9}	& 28.5	& 33.0	& 34.7 \\
\hline
{IIRC-3-CIFAR} & 20.5&  26.0&  	27.1&  	19.6&  	25.6&  25.9&  	16.1&  	35.5 & 	\textbf{37.2}\\
\hline
{IIRC-ImageNet-Subset} & 28.7 & 29.3& 	\textbf{31.7}& 	28.2& 	29.1& 31.3 & 	23.3& 	26.8& 	28.2\\
\hline
\end{tabular}
}
\end{center}

\caption{We show the average of \textit{pw-JS} from comparison over three datasets with and without our HCV module. \textit{+ SPL} means applying HCV in training stage, \textit{+ Infer-HCV} means applying HCV module in inference time.
}
\vspace{-2mm}
\label{tab:3datsets_methods_plus_HCV}
\end{table}

\minisection{HCV applied to existing methods.} To verify the performance of our proposed HCV, we apply it to iCaRL-CNN, iCaRL-norm and LUCIR. 
The average \textit{pw-JS} value is provided in Table~\ref{tab:3datsets_methods_plus_HCV}. We conduct experiments using three different settings, that is IIRC-2-CIFAR, IIRC-3-CIFAR and IIRC-ImageNet-Subset. On IIRC-2-CIFAR setting, with the help of our HCV module during the training stage, the average numbers are increased by nearly 4.3\% for all three different continual learning methods. When we apply HCV also at inference time, it  further improves the consistency of final predictions achieving the average number by 3.2\%, 2.8\%, 1.7\% for these three methods respectively. On the IIRC-3-CIFAR setting, since it is a much harder setup for incremental learning, all these variants suffer a significant drop of performance. LUCIR is much better compared to iCaRL-CNN and iCaRL-norm. Applying HCV in both training and inference stages helps to boost performance around 6.5\% for two iCaRL variants and 21.1\% for LUCIR. IIRC-ImageNet-Subset setting has much higher image diversity, thus it also imposes difficulties for these incremental methods. Under this setting, LUCIR performs worse than iCaRL-CNN and iCaRL-norm even with the improvement from HCV. And iCaRL-CNN works similar to iCaRL-norm but with marginally better performance. Overall, using our proposed HCV during training and inference improves performance of existing methods consistently for different settings. 

\begin{figure*}[tbp!]
\begin{center}
\includegraphics[width=0.99\textwidth]{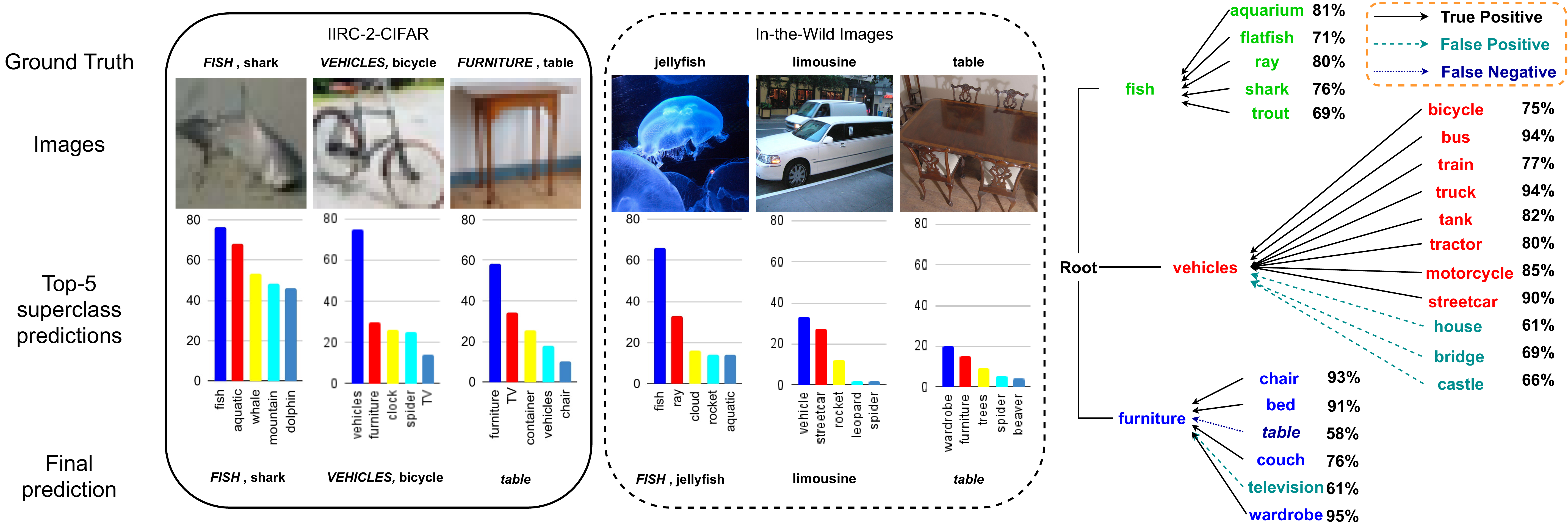}
\end{center}
\vspace{-2mm}
  \caption{{Visual examples of our model applied to IIRC-2-CIFAR setup (annotated with superclasses and subclasses) and in-the-wild images (annotated with class names). We plot the top-5 (ranked by \% percentage) predicted superclasses for each query image. We take the default threshold $\tau=0.6$ to distinguish the success and failure cases. A subgraph of the final predicted graph under IIRC-2-CIFAR setup with iCaRL method is shown on the right. Here the top-1 predicted superclasses with percentages are listed.}
  \vspace{-2mm}
  }
  %\vspace{-5mm}
\label{fig:top5}
\end{figure*}

\minisection{Final estimated hierarchy graph and visual examples.}
After learning the last task under IIRC-2-CIFAR setup when applying our SPL module to iCaRL-CNN, we estimate the full hierarchy and draw a subgraph with 3 superclasses in Fig.~\ref{fig:top5} (right). We can observe that most subclasses are correctly annotated with its superclasses. However \textit{table} is not correctly annotated because its confidence (58\%) does not reach the threshold. 
Interestingly, \textit{television} is wrongly classified as a subclass of \textit{furniture}. In real life, we could also regard it as a member of \textit{furniture} and this was learned because \textit{televisions} occur often in \textit{furniture} scenes. This kind of information can help human operators in annotating and verifying the dataset hierarchy. Further, we see that \textit{house, bridge, castle} are false positives, and are classified as subclasses of \textit{vehicles}. This could be because \textit{vehicles} images co-occur with the \textit{house, bridge, castle} classes as their background. Finally, we also show some visual examples from IIRC-2-CIFAR setup and in-the-wild images in Fig.~\ref{fig:top5}(left).

\begin{figure*}[tb]

\begin{minipage}[b]{0.31\linewidth}
\centering
\includegraphics[width=\textwidth]{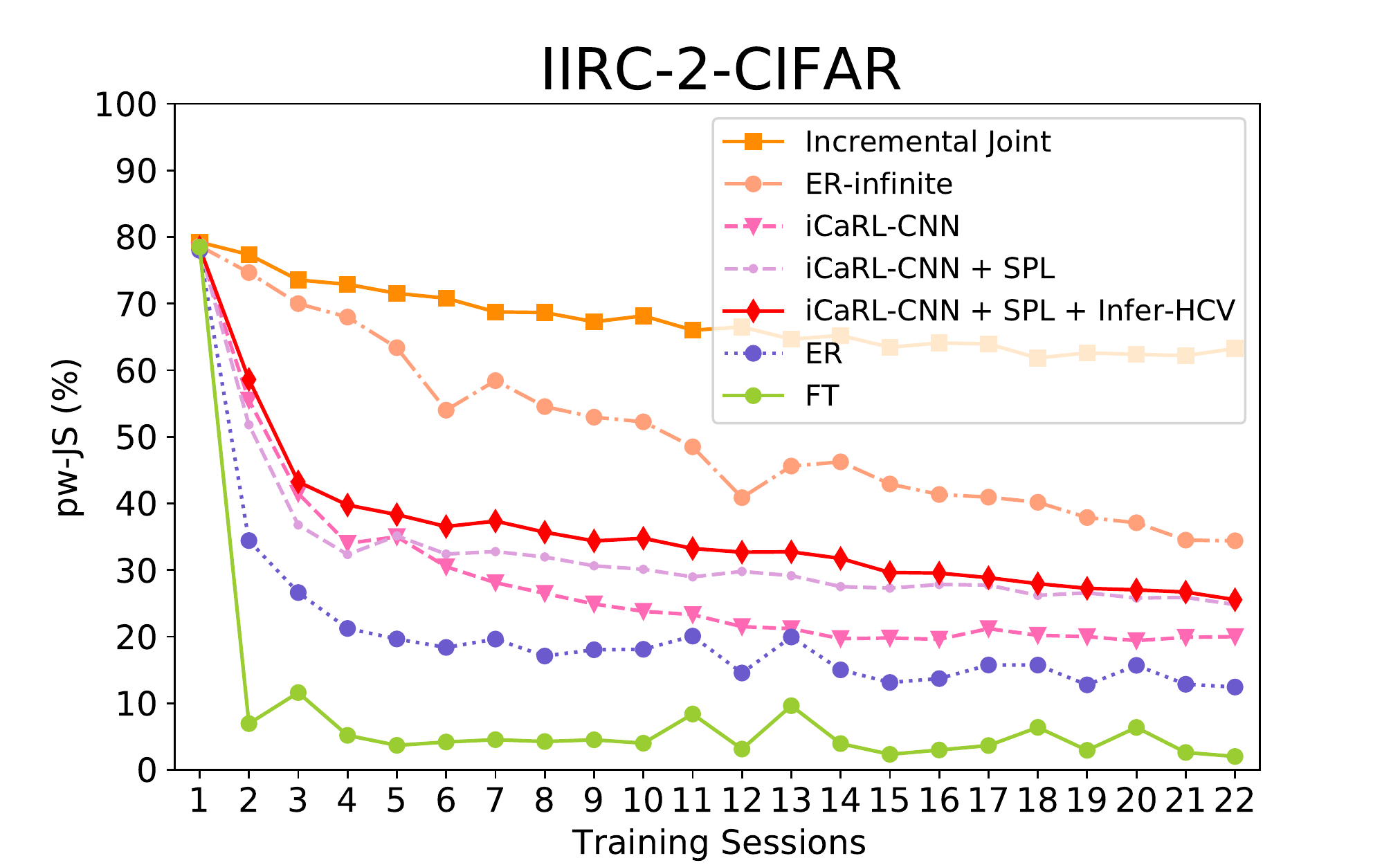}
% \subcaption{iCaRL-CNN}
\label{fig:iirc_2_cifar_22task_icarl_cnn}
\end{minipage}
\begin{minipage}[b]{0.31\linewidth}
\centering
\includegraphics[width=\textwidth]{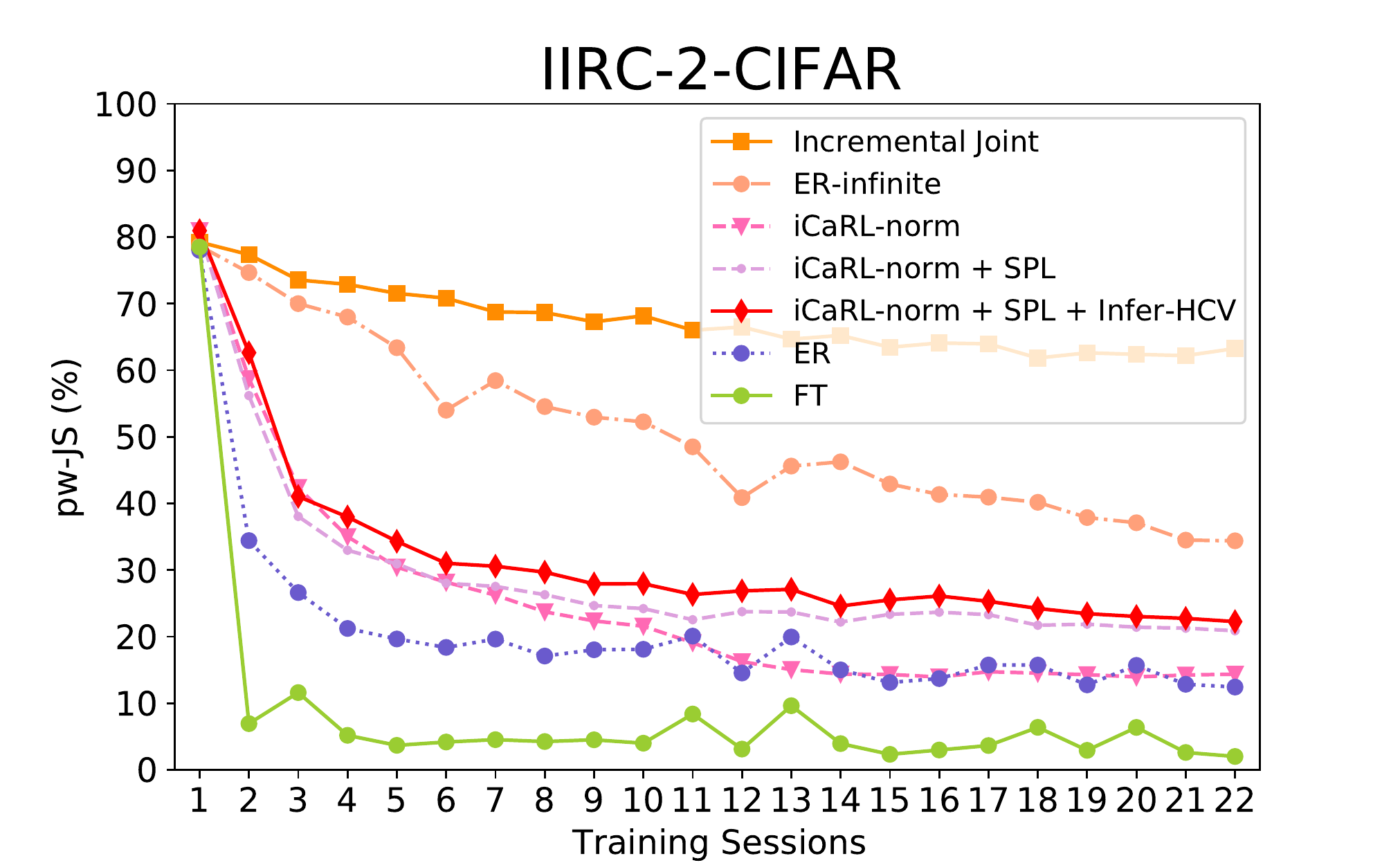}
% \subcaption{iCaRL-norm}
\label{fig:iirc_2_cifar_22task_icarl_norm}
\end{minipage}
\begin{minipage}[b]{0.31\linewidth}
\centering
\includegraphics[width=\textwidth]{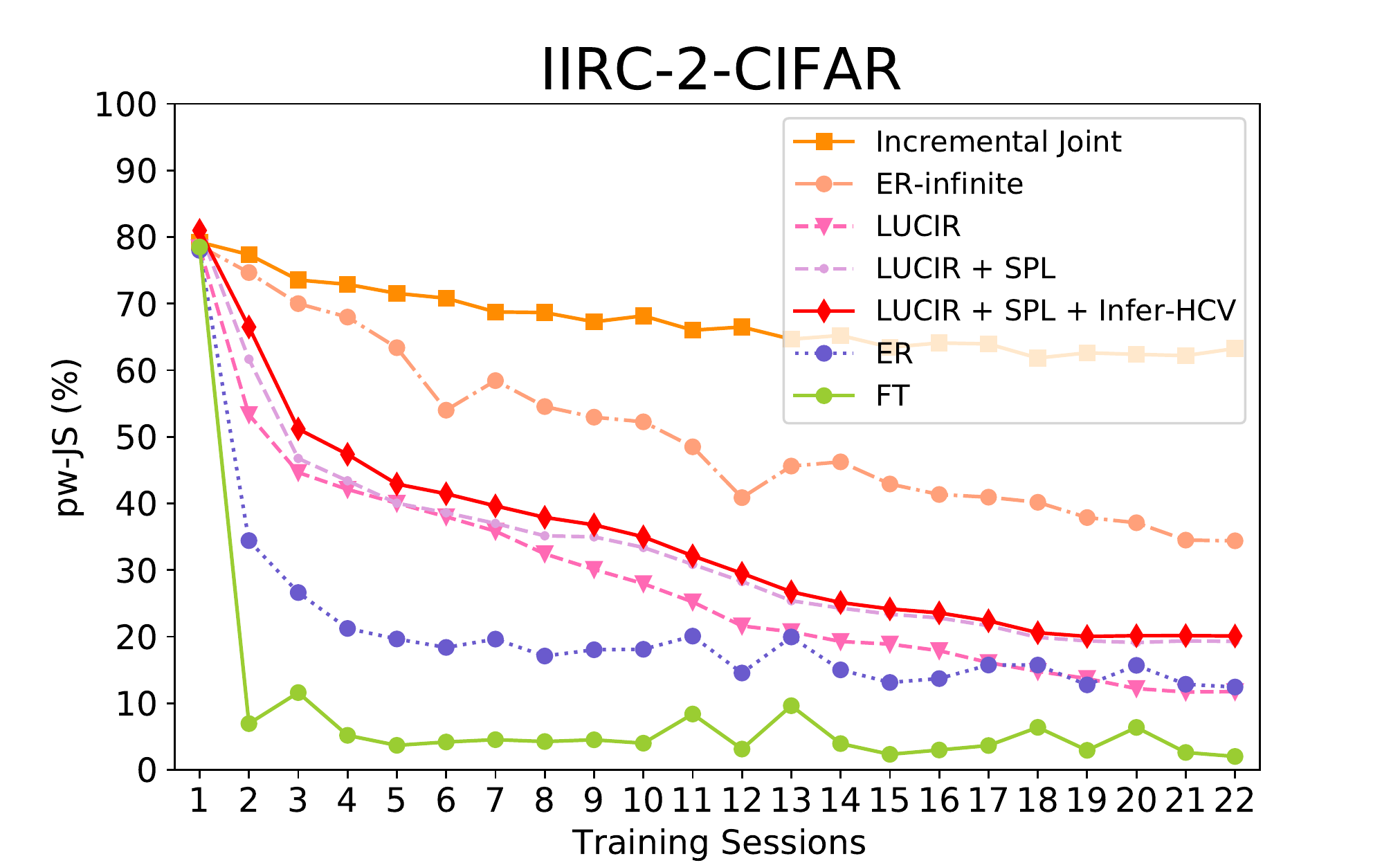}
% \subcaption{LUCIR}
\label{fig:iirc_2_cifar_22task_lucir}
\end{minipage}

\begin{minipage}[b]{0.31\linewidth}
\centering
\includegraphics[width=\textwidth]{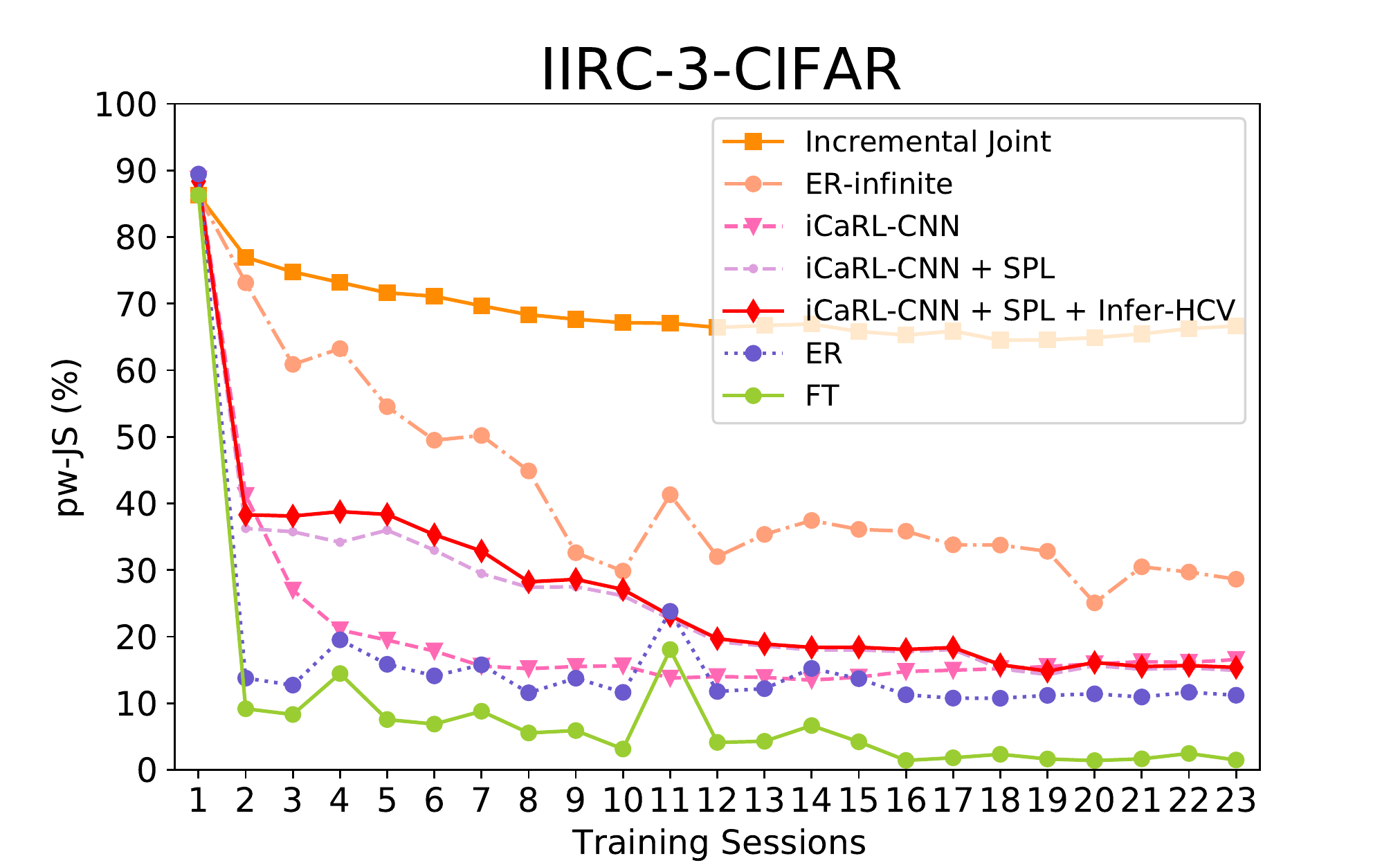}
% \subcaption{iCaRL-CNN}
\label{fig:iirc_3_cifar_23task_icarl_cnn}
\end{minipage}
\begin{minipage}[b]{0.31\linewidth}
\centering
\includegraphics[width=\textwidth]{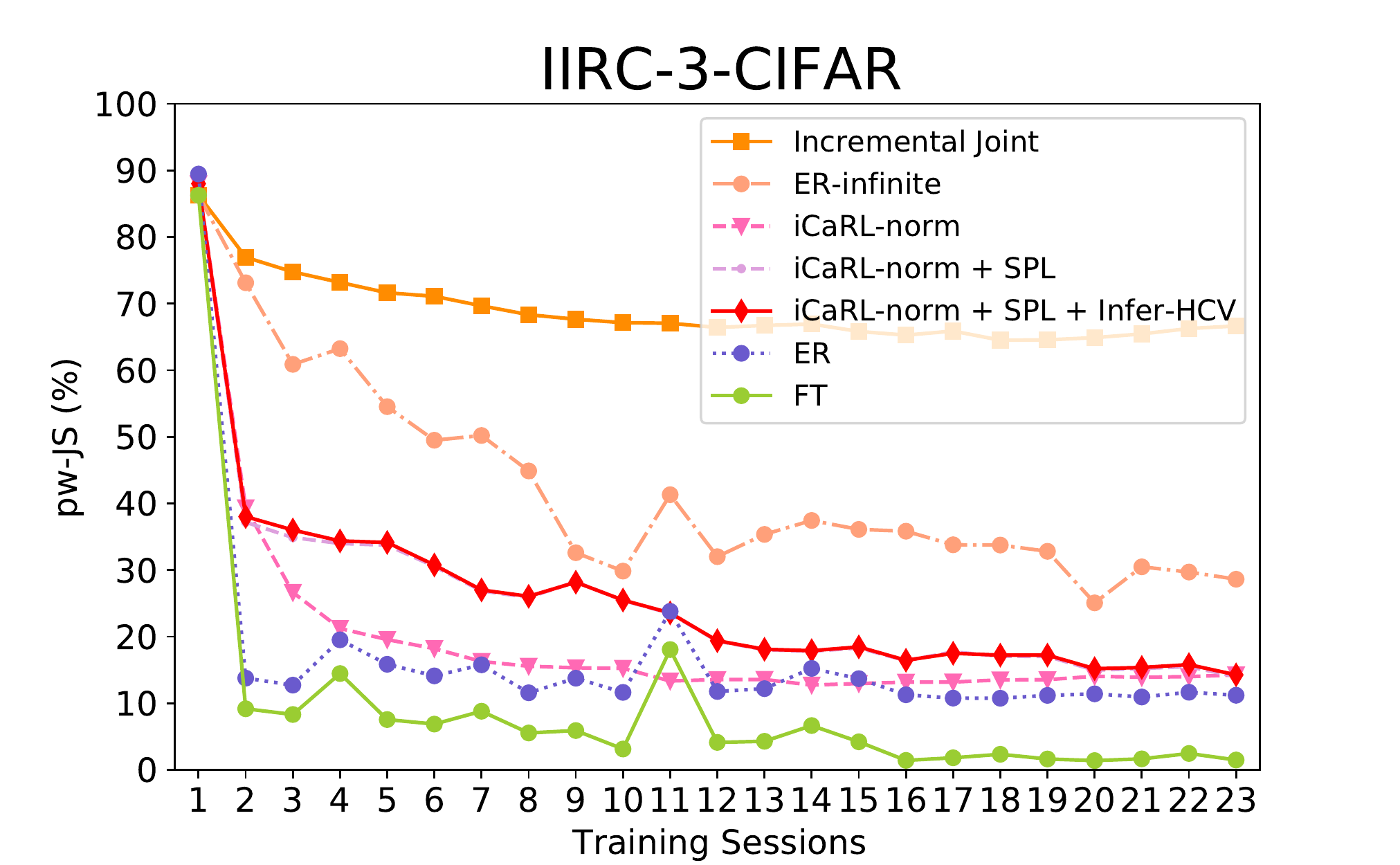}
% \subcaption{iCaRL-norm}
\label{fig:iirc_3_cifar_23task_icarl_norm}
\end{minipage}
\begin{minipage}[b]{0.31\linewidth}
\centering
\includegraphics[width=\textwidth]{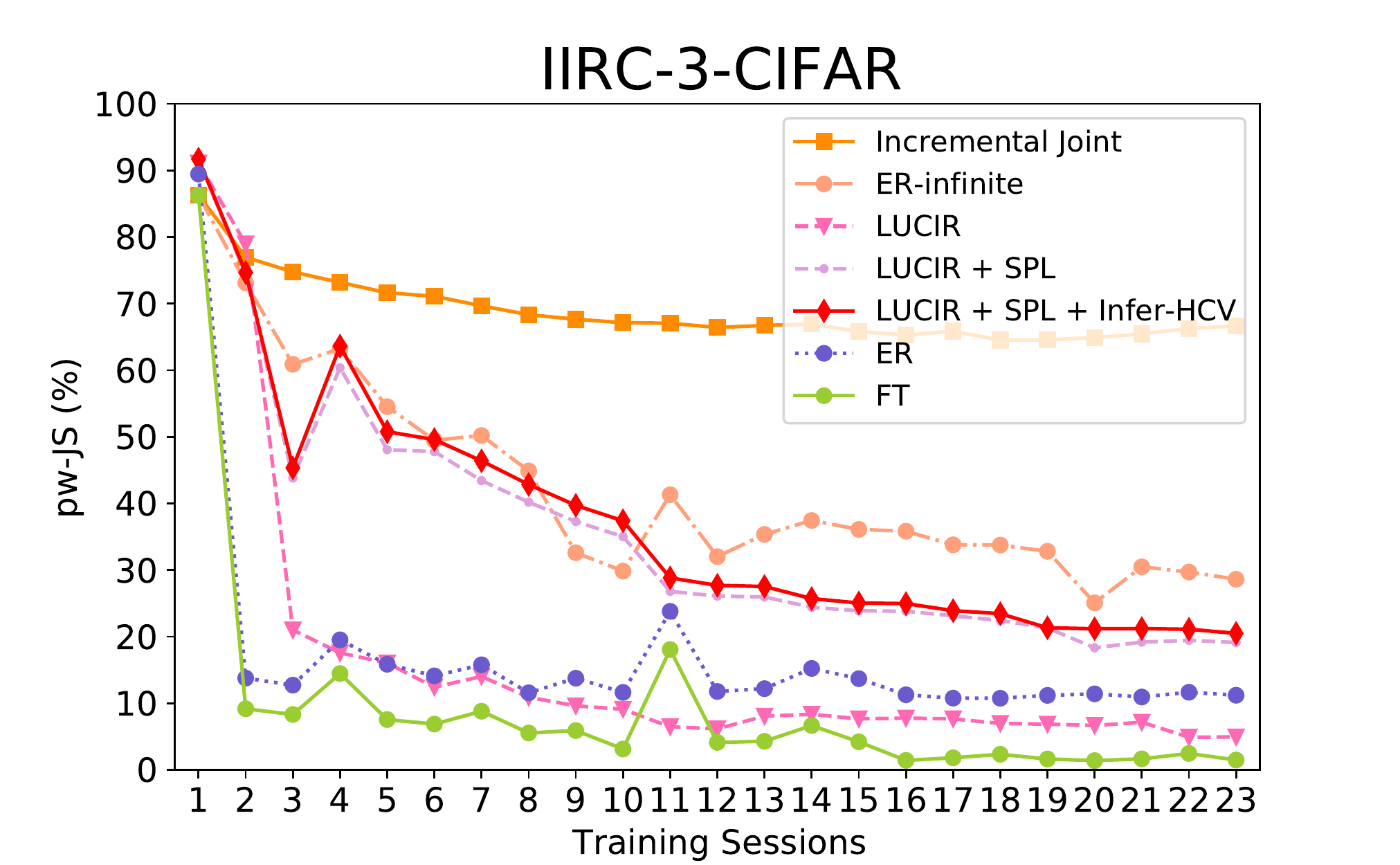}
% \subcaption{LUCIR}
\label{fig:iirc_3_cifar_23task_lucir}
\end{minipage}

\begin{minipage}[b]{0.31\linewidth}
\centering
\includegraphics[width=\textwidth]{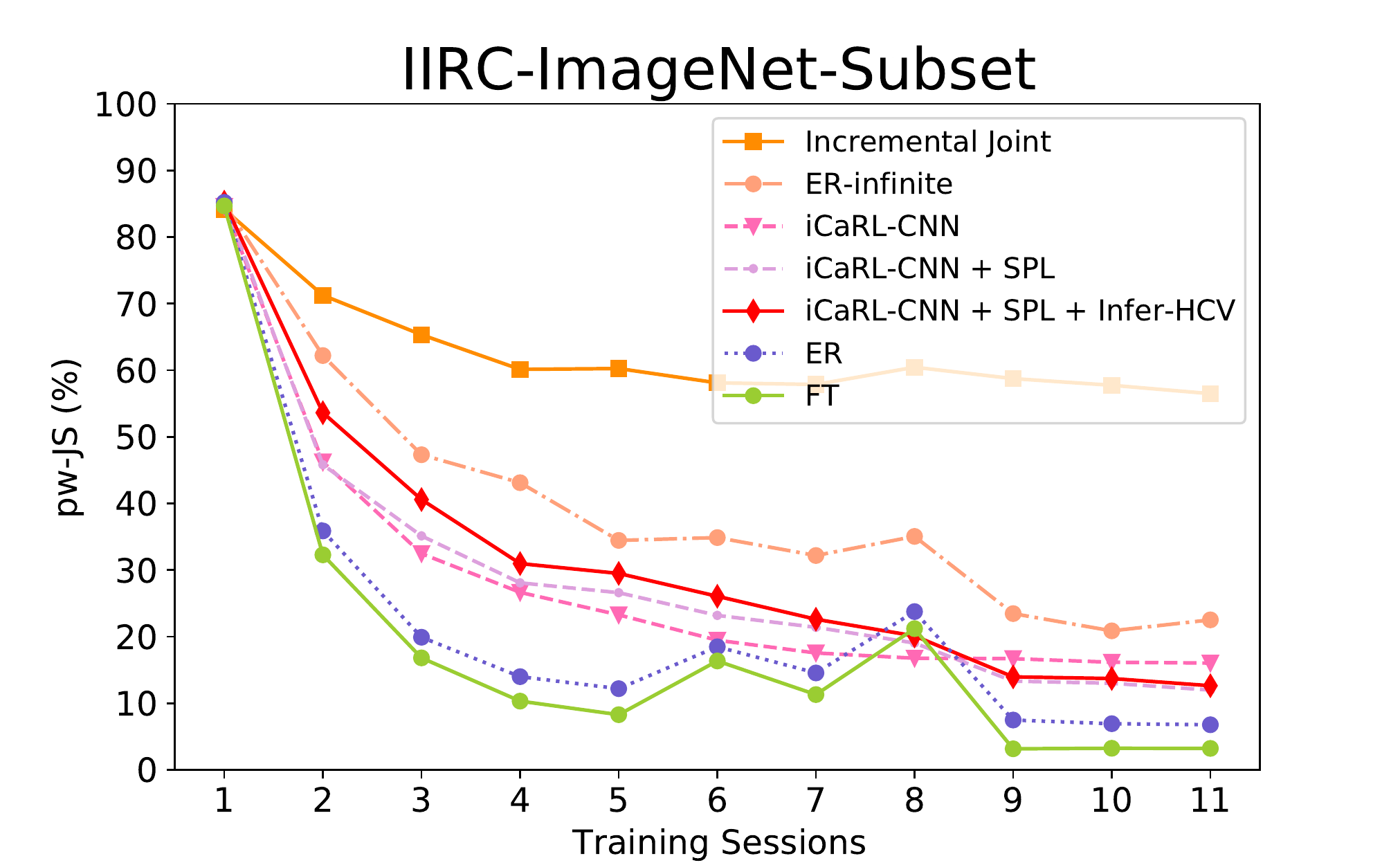}
\subcaption{iCaRL-CNN}
\label{fig:ImageNet_Subset_11task_icarl_cnn}
\end{minipage}
\begin{minipage}[b]{0.31\linewidth}
\centering
\includegraphics[width=\textwidth]{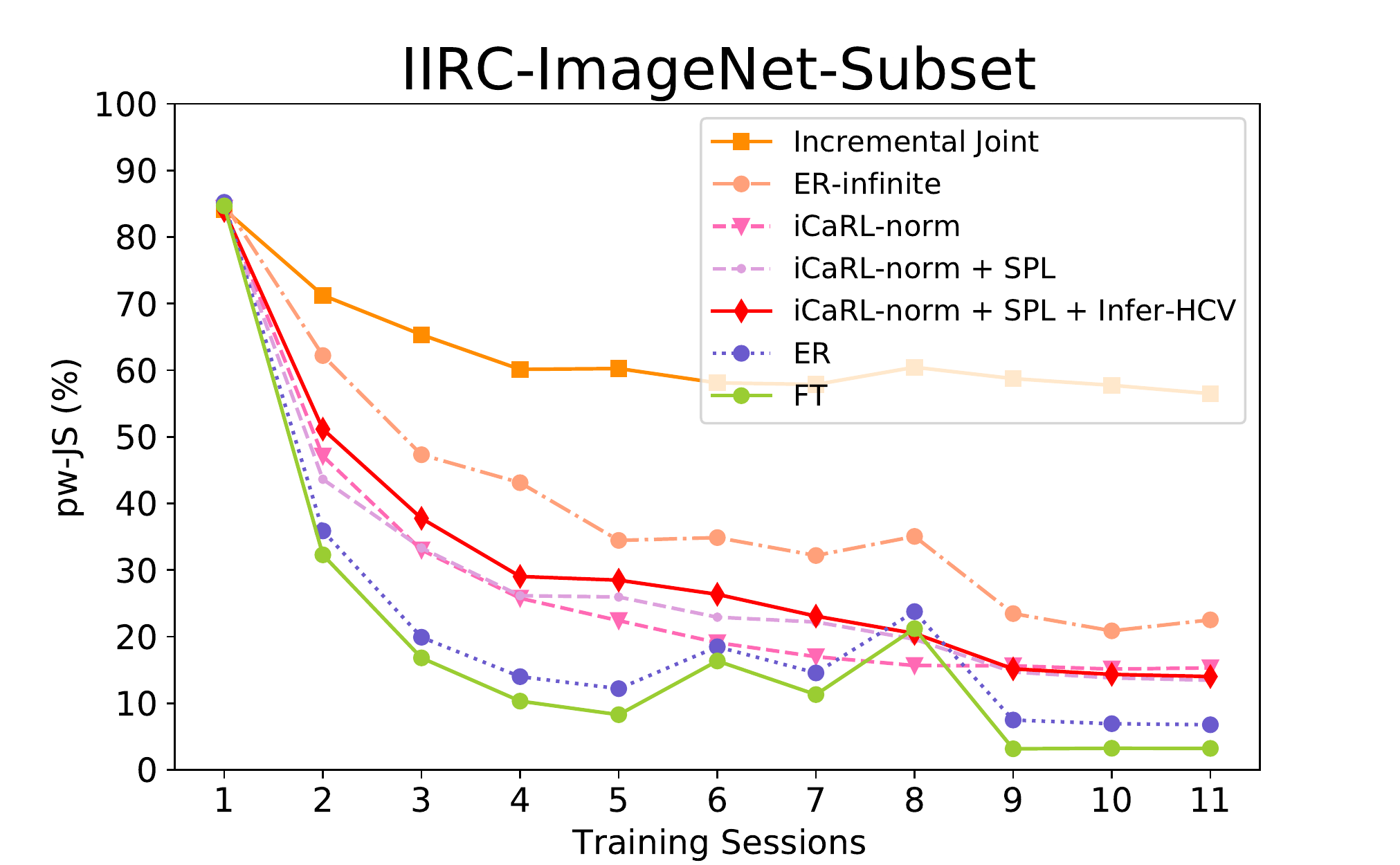}
\subcaption{iCaRL-norm}
\label{fig:ImageNet_Subset_11task_icarl_norm}
\end{minipage}
\begin{minipage}[b]{0.31\linewidth}
\centering
\includegraphics[width=\textwidth]{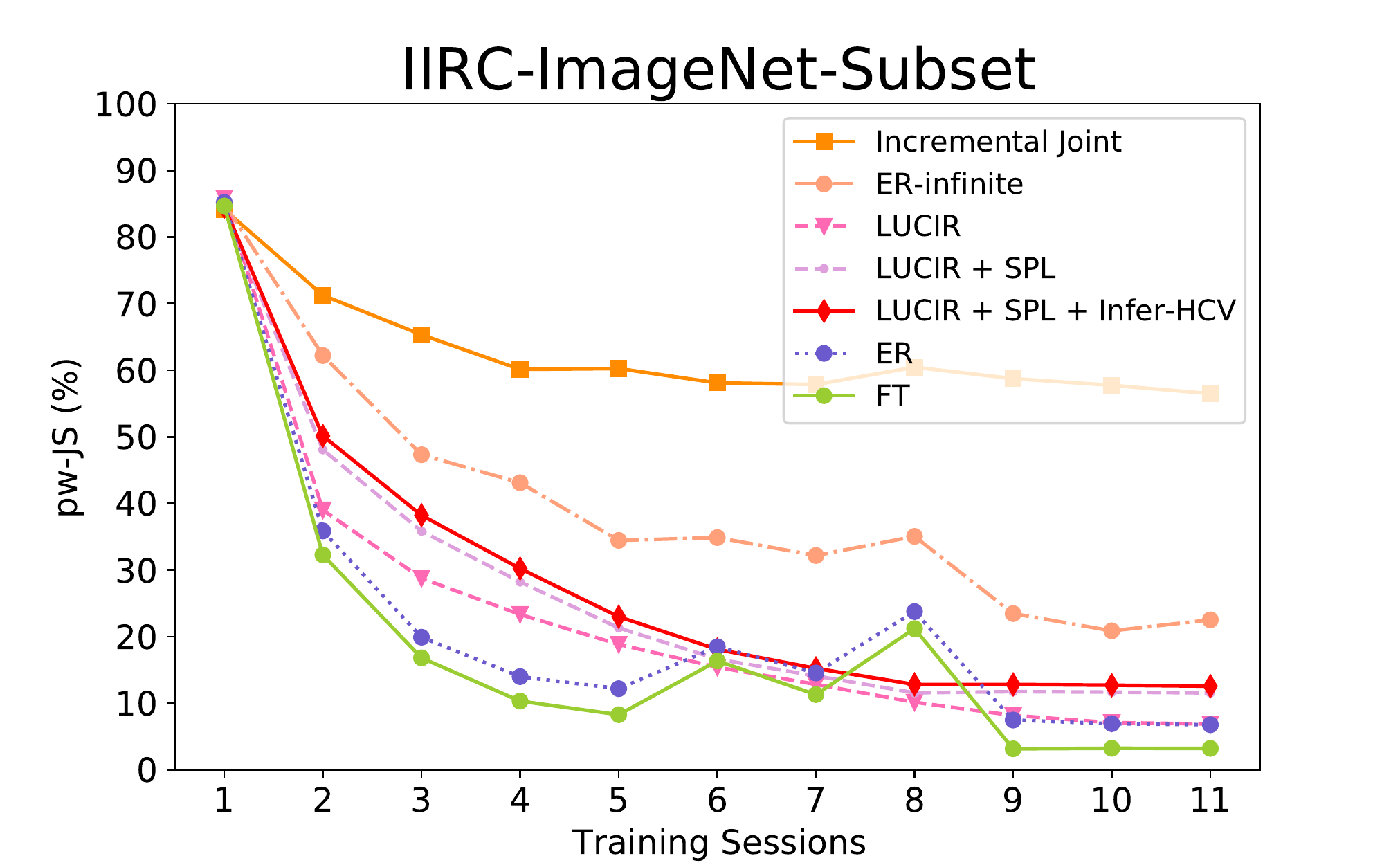}
\subcaption{LUCIR}
\label{fig:ImageNet_Subset_11task_lucir}
\end{minipage}

\vspace{-2mm}
\caption{Experimental results over IIRC-2-CIFAR, IIRC-3-CIFAR and IIRC-ImageNet-Subset setups based on three methods: iCaRL-CNN, iCaRL-norm and LUCIR.
\vspace{-2mm}
}
\vspace{-2mm}
\label{fig:HCV_expr}
\end{figure*}

\minisection{Comparison with SOTA methods.}
In Fig.~\ref{fig:HCV_expr} we plot the dynamic performance changes of different methods. The general trend on different settings are similar. Incremental Joint always achieves the best results as an upper bound, benefiting from access to all data and labels, while ER-infinite lacks the knowledge of full labels resulting in a worse performance. Our proposed HCV improves existing methods consistently, but the gap between our best and the two upper bounds (ER-infinite and Incremental Joint) is still large, which shows that IIRC setting is a very challenging setting requiring more research.

\begin{figure*}[tb]
\begin{minipage}[b]{0.20\linewidth}
\centering
\includegraphics[width=\textwidth]{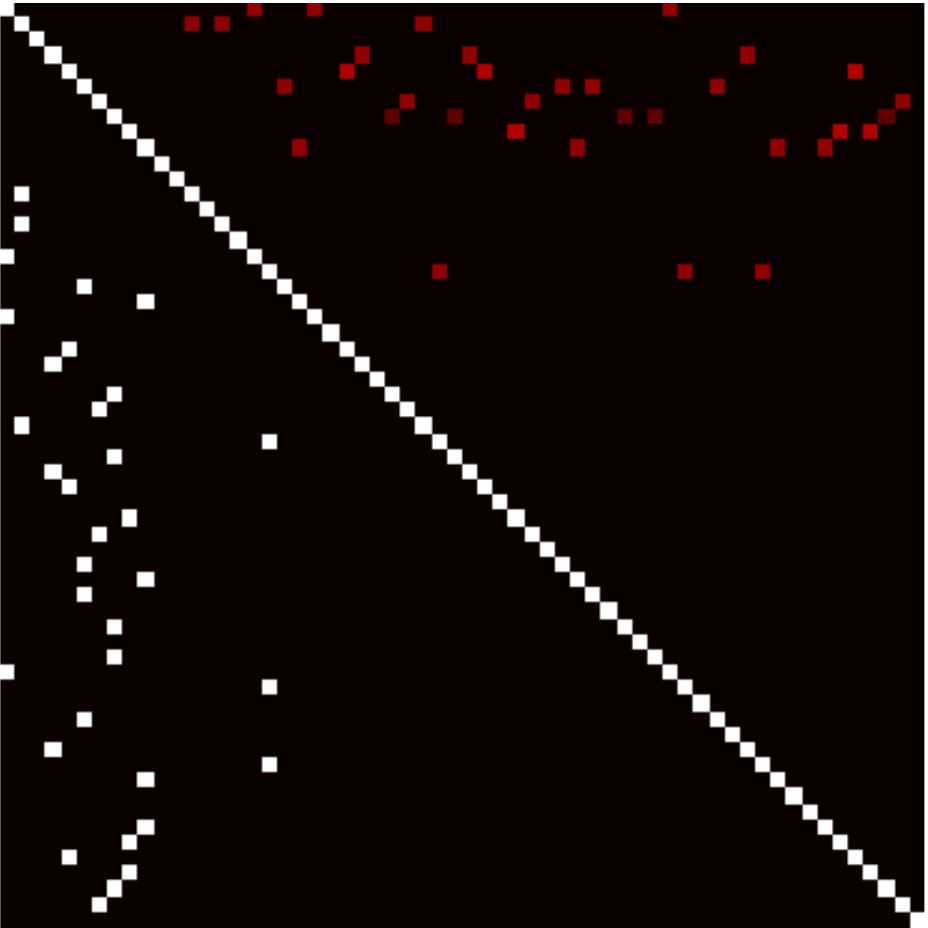}
\subcaption{Ground truth}
\label{fig:2_GT_confusion_matrix_10_icarl_cnn_ptm}
\end{minipage}
\hspace{5mm}
\begin{minipage}[b]{0.20\linewidth}
\centering
\includegraphics[width=\textwidth]{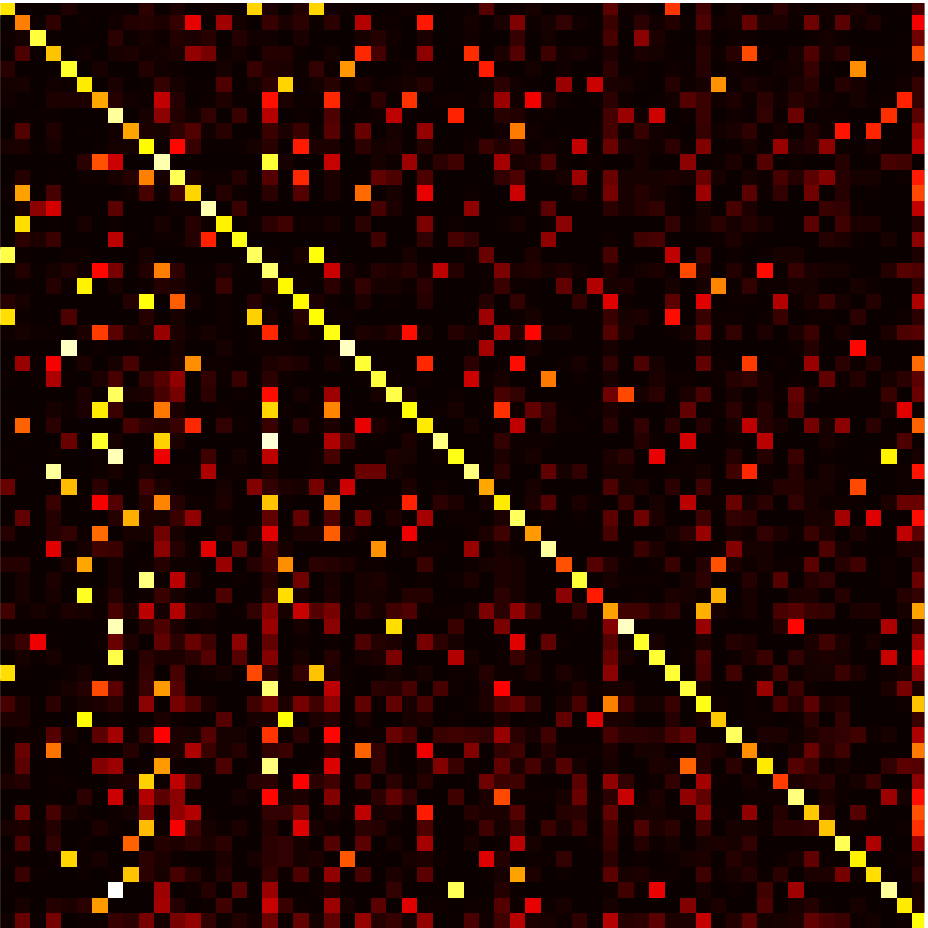}
\subcaption{iCaRL-CNN}
\label{fig:2_ori_confusion_matrix_10_icarl_cnn_ptm}
\end{minipage}
\hspace{5mm}
\begin{minipage}[b]{0.20\linewidth}
\centering
\includegraphics[width=\textwidth]{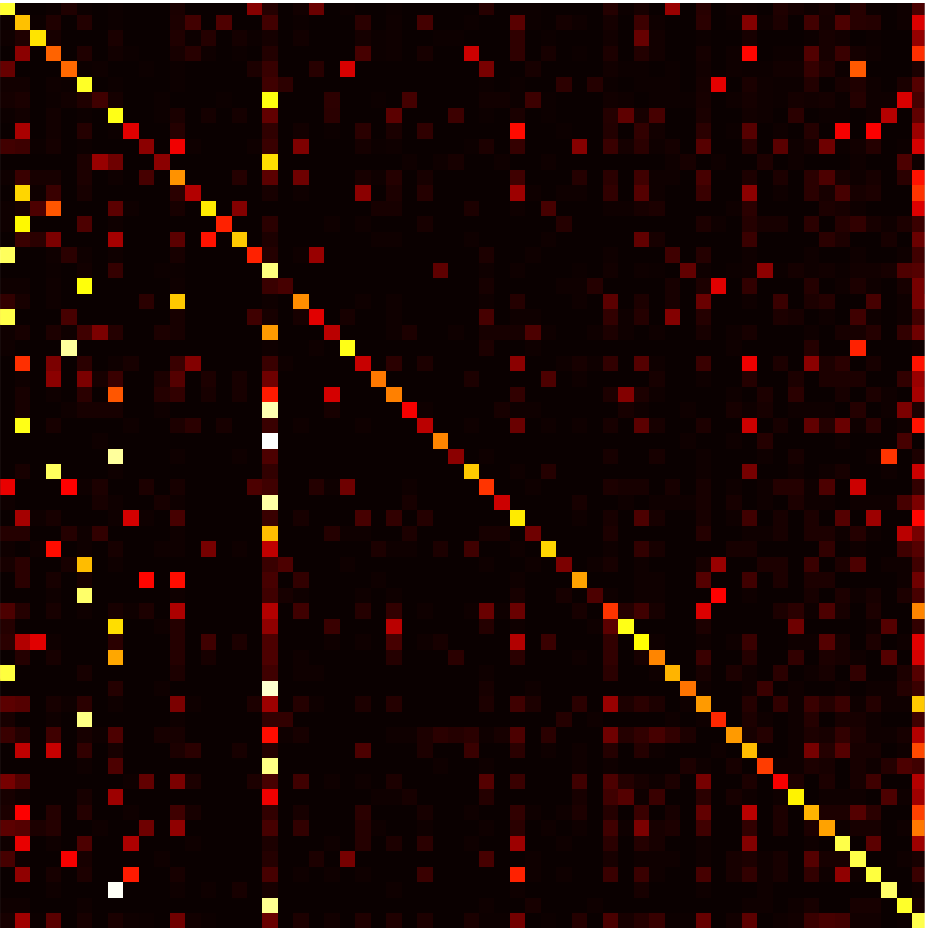}
\subcaption{+ SPL}
\label{fig:2_train_confusion_matrix_10_icarl_cnn_ptm}
\end{minipage}
\hspace{5mm}
\begin{minipage}[b]{0.20\linewidth}
\centering
\includegraphics[width=\textwidth]{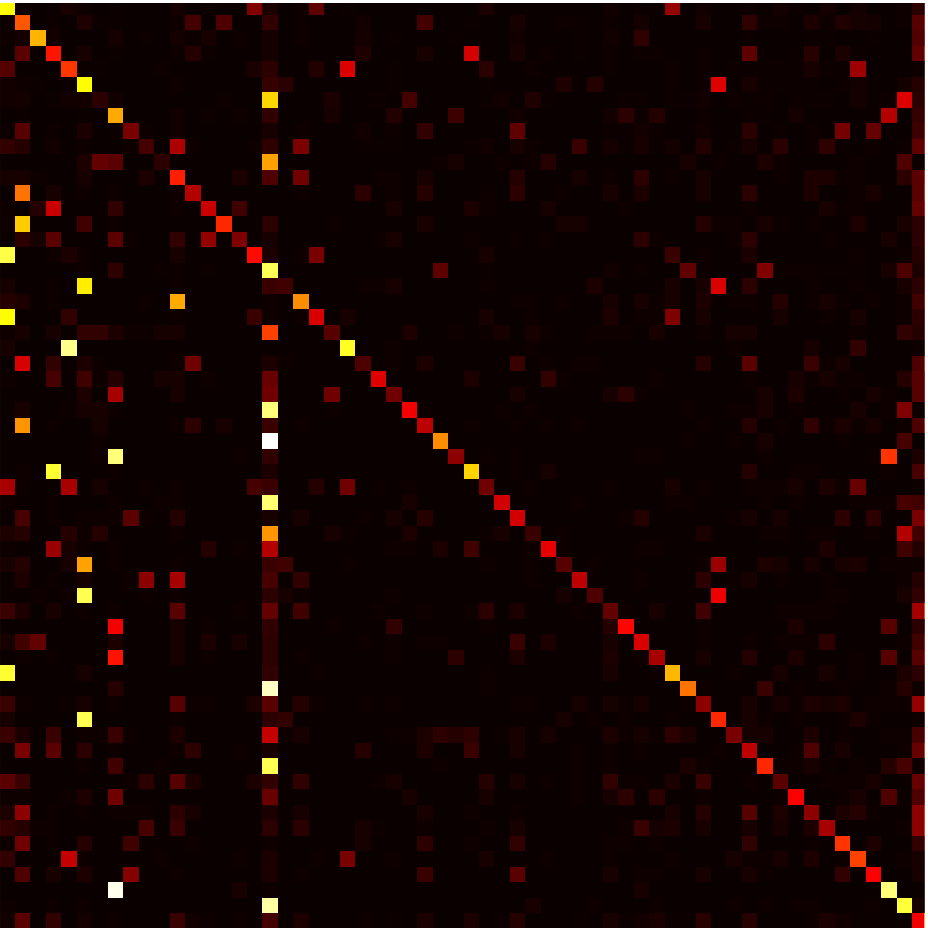}
\subcaption{+SPL+HCV}
\label{fig:2_both_confusion_matrix_10_icarl_cnn_ptm}
\end{minipage}

\vspace{-2mm}
\caption{Confusion matrices of groundtruth, original continual learning methods, applying SPL and applying Infer-HCV to iCaRL-CNN after task 11 under IIRC-2-CIFAR setup. 
\vspace{-2mm}
}
\vspace{-2mm}
\label{fig:confusion_matrices}
\end{figure*}

\minisection{Confusion matrices.} Fig.~\ref{fig:confusion_matrices} shows the confusion matrices after learning task 11 under IIRC-2-CIFAR setup. They are from the ground truth, original continual learning methods, and HCV applied to both training and inference time. It can be observed that after using HCV, the redundant predictions are cleaned with our learned prior knowledge about the classes hierarchy, therefore HCV plays a role of a de-noising procedure for confusion matrices. 

\subsection{Ablation study}
\begin{figure*}[tb]
\begin{minipage}[b]{0.325\linewidth}
\centering
\includegraphics[width=\textwidth]{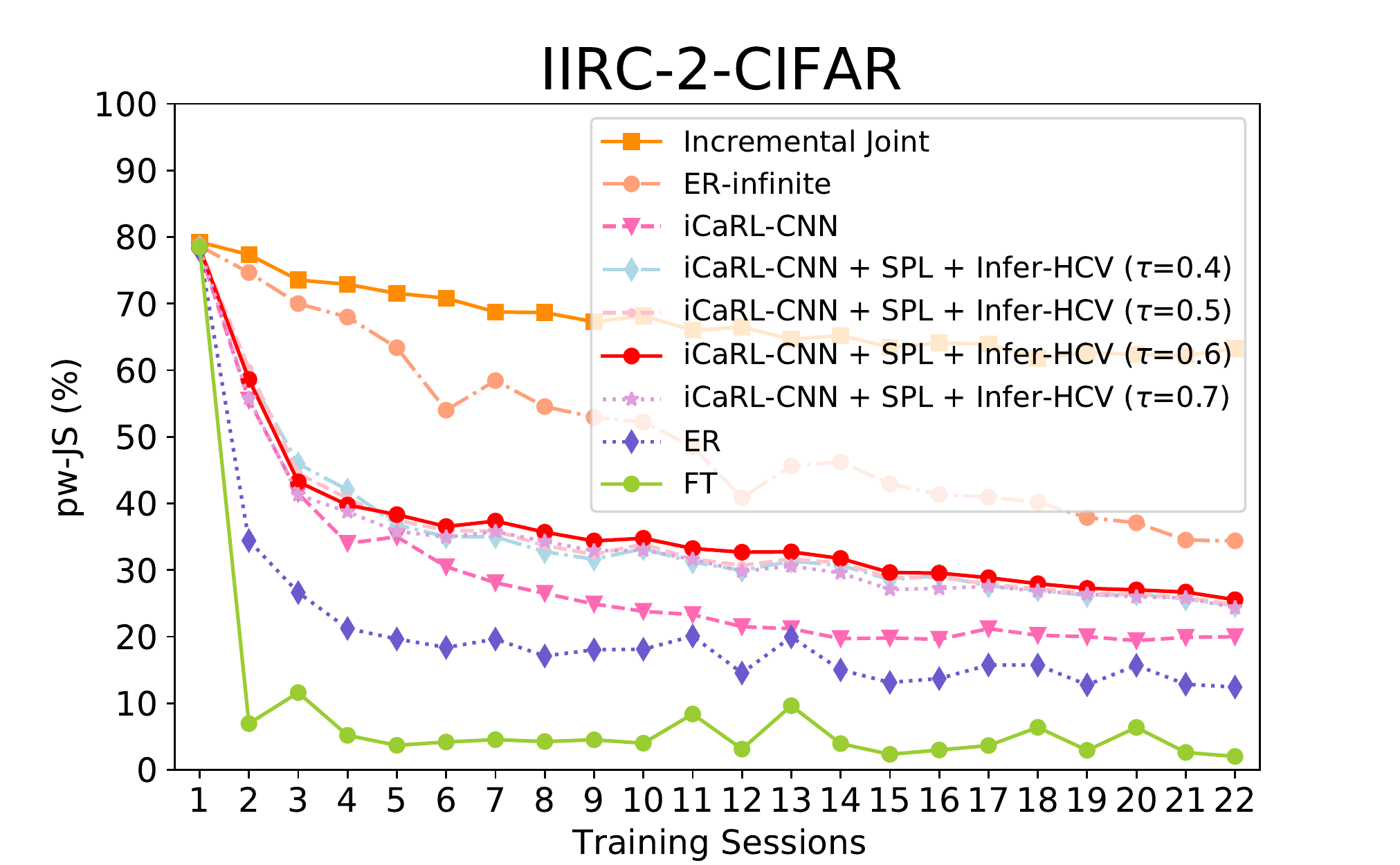}
\subcaption{Ablation on threshold $\tau$}
\label{fig:ablate_th_curve}
\end{minipage}
\begin{minipage}[b]{0.325\linewidth}
\centering
\includegraphics[width=\textwidth]{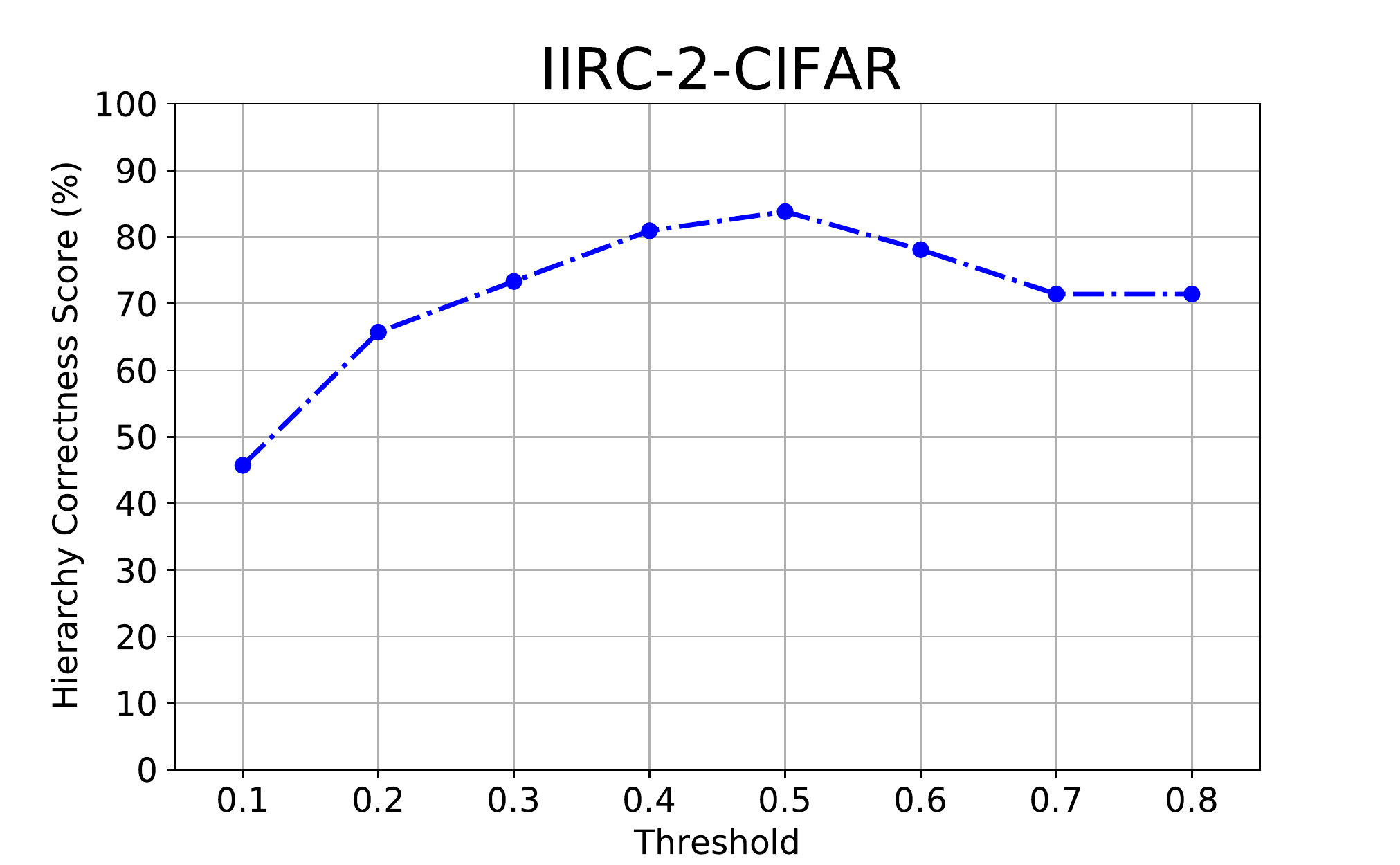}
\subcaption{HCS changes by $\tau$}
\label{fig:ablate_th_rate}
\end{minipage}
\begin{minipage}[b]{0.325\linewidth}
\centering
\includegraphics[width=\textwidth]{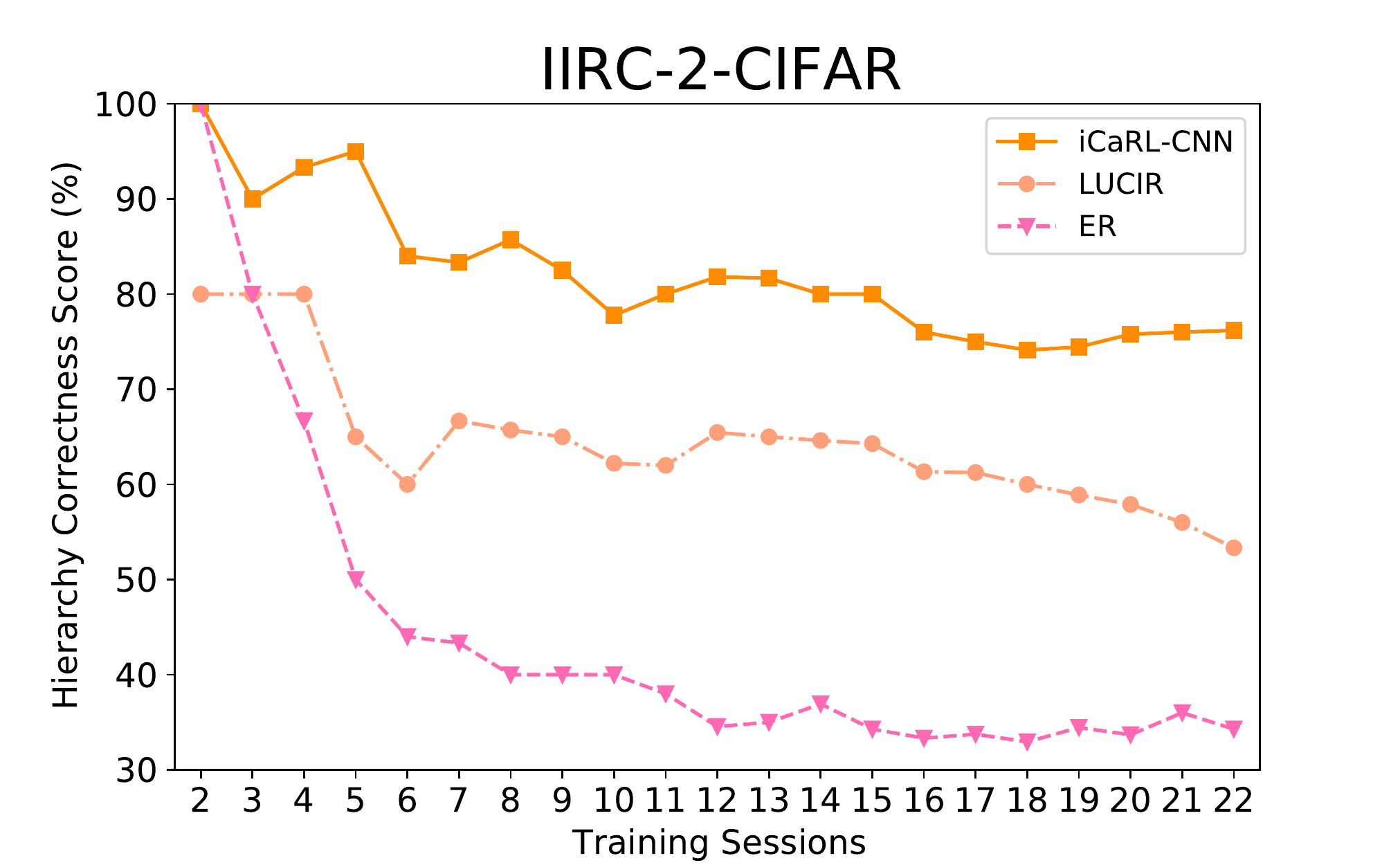}
\subcaption{HCS curves of 3 methods}
\label{fig:rebuttal_iirc_2_cifar_22task_hierarchy_acc}
\end{minipage}

\begin{minipage}[b]{0.325\linewidth}
\centering
\includegraphics[width=\textwidth]{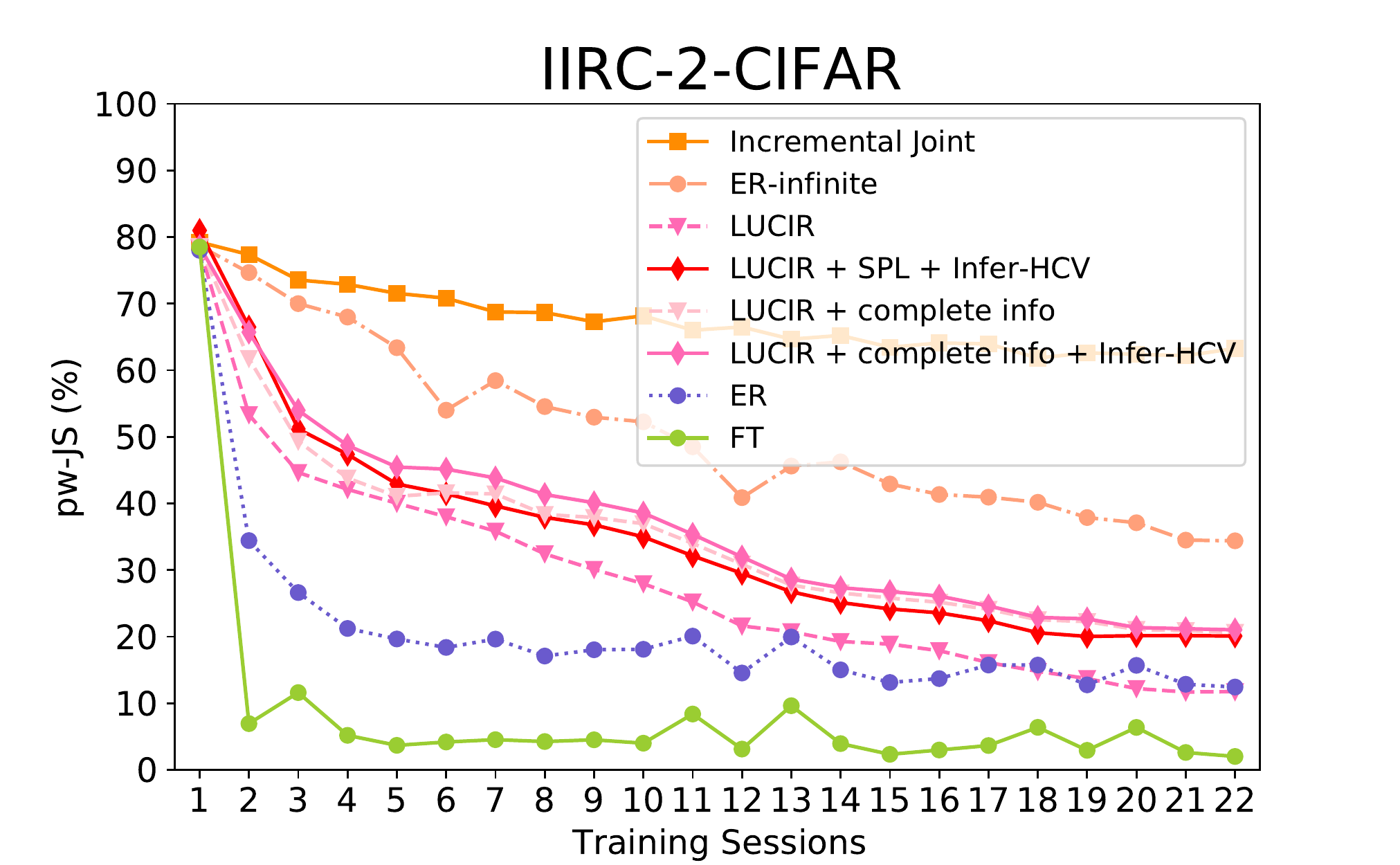}
\subcaption{Ablate HCS on LUCIR}
\label{fig:rebuttal_iirc_2_cifar_22task_LUCIR_complete_info}
\end{minipage}
\begin{minipage}[b]{0.325\linewidth}
\centering
\includegraphics[width=\textwidth]{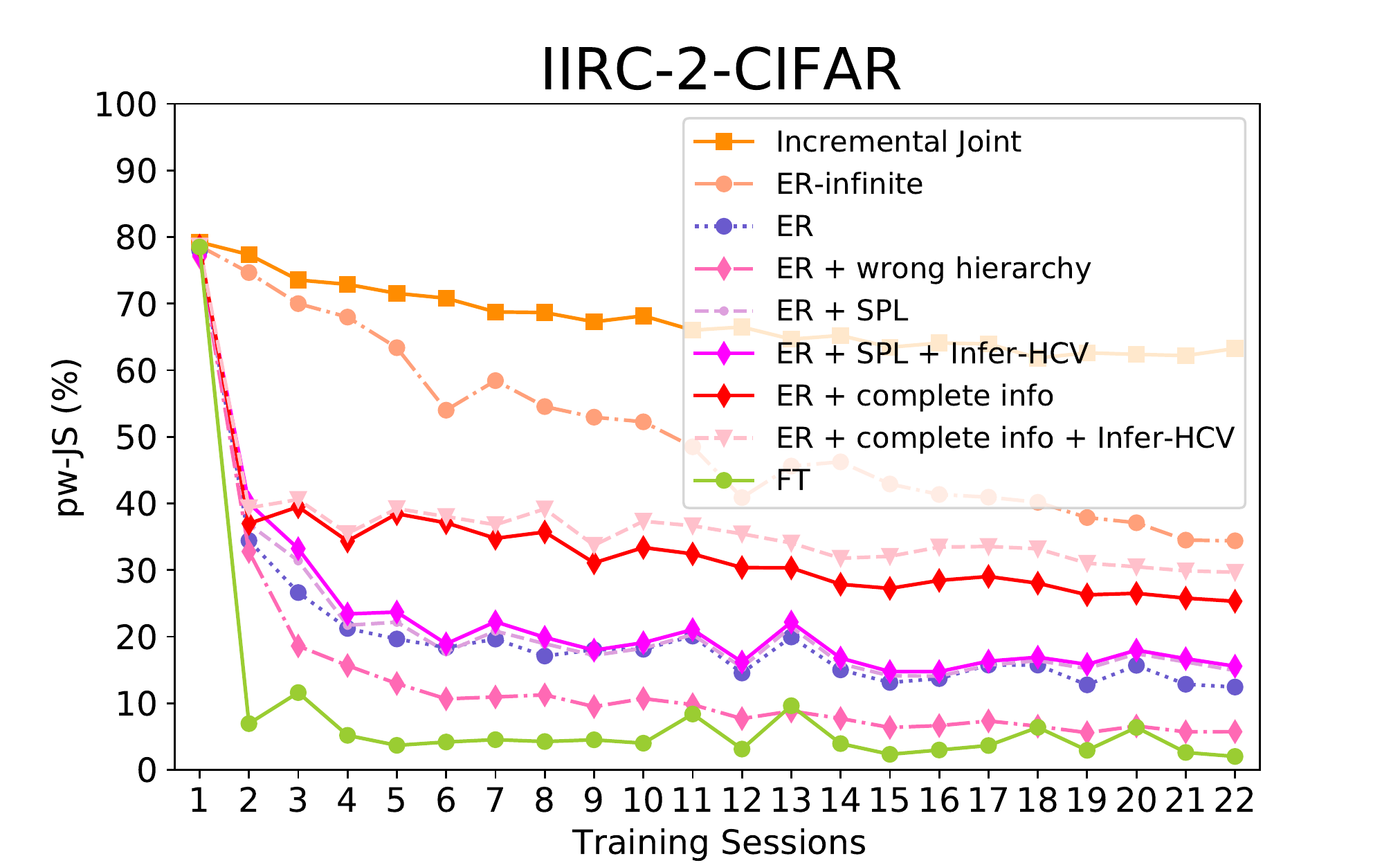}
\subcaption{Ablate HCS on ER}
\label{fig:rebuttal_iirc_2_cifar_22task_ER}
\end{minipage}
\begin{minipage}[b]{0.325\linewidth}
\centering
\includegraphics[width=\textwidth]{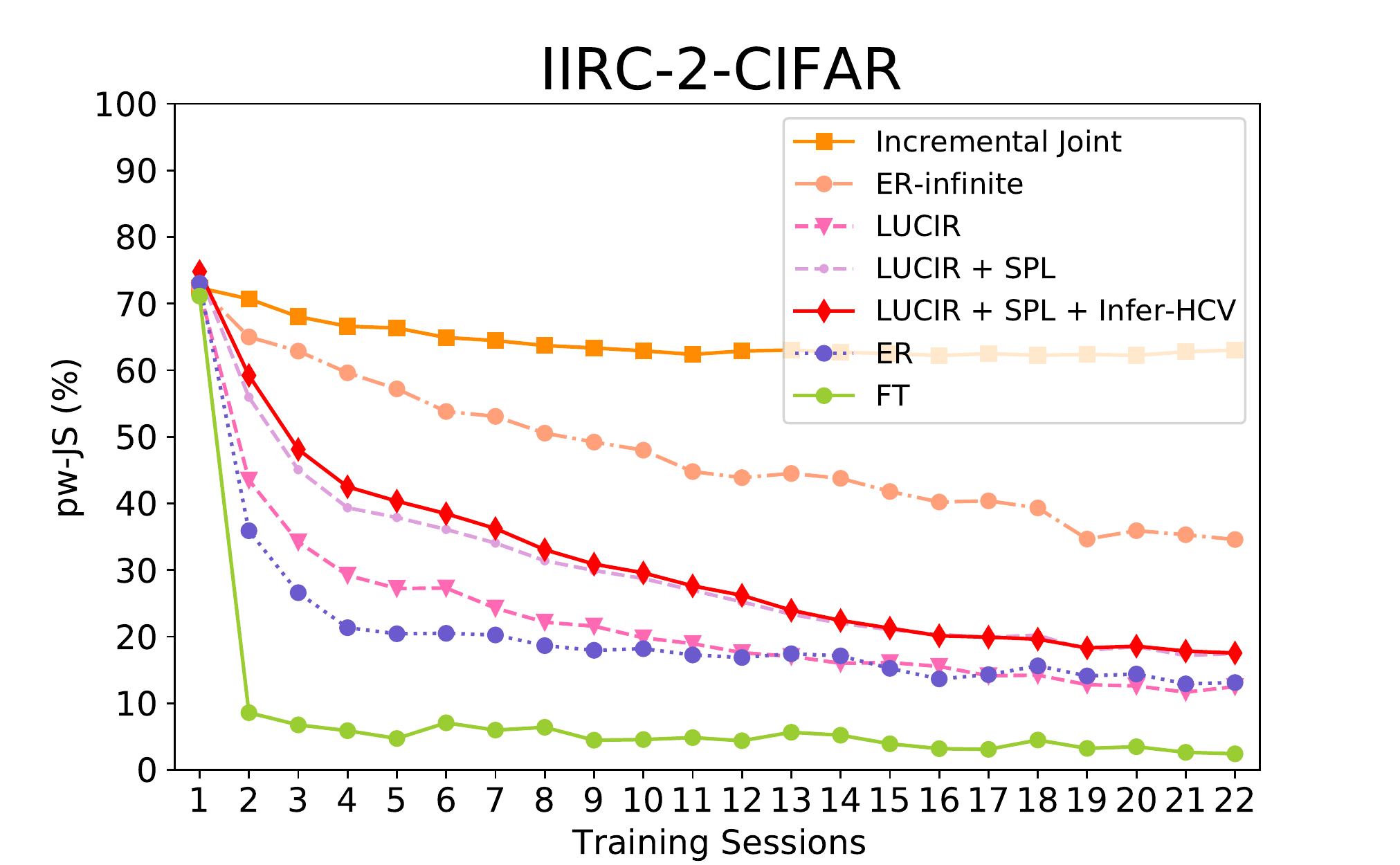}
\subcaption{LUCIR with 10 orders}
\label{fig:rebuttal_iirc_2_cifar_22task_lucir}
\end{minipage}
% \vspace{-2mm}
\caption{{
Ablation study over threshold $\tau$, HCS and class orders on IIRC-2-CIFAR setup. }
% \vspace{-2mm}
}
% \vspace{-2mm}
\label{fig:ablation_study_over_correct_rate}
\end{figure*}

\minisection{Ablation study over threshold $\tau$.}
We conduct an ablation study on the threshold $\tau$ under IIRC-2-CIFAR setup. In Fig.~\ref{fig:ablate_th_curve}, we compare the values of $\tau$ \{0.4, 0.5, 0.6, 0.7\} when applying HCV on both training and inference stages. We can observe that with different hyper-parameters, it improves over iCaRL-CNN consistently. In Fig.~\ref{fig:ablate_th_rate}, we show how the hierarchy correctness score (HCS) changes with the threshold from 0.1 to 0.8, and is around 75\% to 80\% when $\tau$ is in the range [0.3, 0.7]. In our experiments, we set $\tau=0.6$ by default.

\minisection{Ablation study over hierarchy correctness score (HCS).}
We also conduct an ablation study over the HCS on LUCIR and ER methods as shown in Fig.~\ref{fig:rebuttal_iirc_2_cifar_22task_LUCIR_complete_info} and Fig.~\ref{fig:rebuttal_iirc_2_cifar_22task_ER}. The hierarchy correctness scores for iCaRL, LUCIR, ER are 76.2\%, 56.0\%, 34.3\%, respectively (the HCS curves by training sessions are shown in Fig.~\ref{fig:rebuttal_iirc_2_cifar_22task_hierarchy_acc}). The higher hierarchy correctness score for iCaRL-CNN helps it achieve  state-of-the-art performance on IIRC-2-CIFAR and IIRC-ImageNet-Subset (Table~\ref{tab:3datsets_methods_plus_HCV} and Fig.~\ref{fig:HCV_expr}). While LUCIR achieves a much lower score though it is regarded as one of the best methods in continual learning~\cite{masana2020class}.

We also show the performance of the LUCIR and ER methods with the ground-truth hierarchy, which means it has a HCS of 100\% (see Fig.~\ref{fig:rebuttal_iirc_2_cifar_22task_LUCIR_complete_info} and Fig.~\ref{fig:rebuttal_iirc_2_cifar_22task_ER}). In this case 3.0\% and 15.0\% improvements are observed for LUCIR and ER respectively. That implies that our HCV module can benefit from a preciser hierarchy estimation to reduce the gap to ER-infinite. To test how a completely wrong class hierarchy influences our model, we randomly generate a hierarchy for IIRC-2-CIFAR and apply it to ER (Fig.~\ref{fig:rebuttal_iirc_2_cifar_22task_ER}), we can observe a drop of HCS from 34.3\% to 0.0\%, and the overall performance drops for ER to nearly 7.0\%.

\minisection{HCV (on LUCIR) performance with 10 orders.}
In Fig.~\ref{fig:rebuttal_iirc_2_cifar_22task_lucir} the experiments are conducted with all 10 task-orderings proposed in IIRC~\cite{abdelsalam2020iirc}. We plot the average performance. Here we apply our SPL and Infer-HCV to the LUCIR model. We observe a significant and consistent improvement compared to the ER baseline ($\approx$10.0\%) and the basic LUCIR method ($\approx$8.0\%). In conclusion, our method improves the performance under various orders and settings.

\section{Conclusion}

In this paper, we proposed a Hierarchy-Consistency Verification module for Incremental Implicit-Refined Classification (IIRC) problem. With this module, we can boost the existing incremental learning methods by a large margin. From our experiments on three different setups, we evaluate and prove the effectiveness of our proposed module during both training and inference. And from the visualization of confusion matrices, we can also find that our HCV module works as a denoising method to the confusion matrices.
For future work, we are interested in associating hierarchical classification, multi-label classification with IIRC problem, thus to have a more robust model to overcome forgetting in more realistic setups.

\section*{Acknowledgement}
We acknowledge the support from Huawei Kirin Solution, the Spanish Government funding for projects PID2019-104174GB-I00 and RTI2018-102285-A-I00, and Kai Wang acknowledges the Chinese Scholarship Council (CSC) No.201706170035. Herranz acknowledges the Ramón y Cajal fellowship RYC2019-027020-I.

\bibliography{longstrings,egbib}
\includepdf[pages=-,pagecommand={},width=\textwidth]{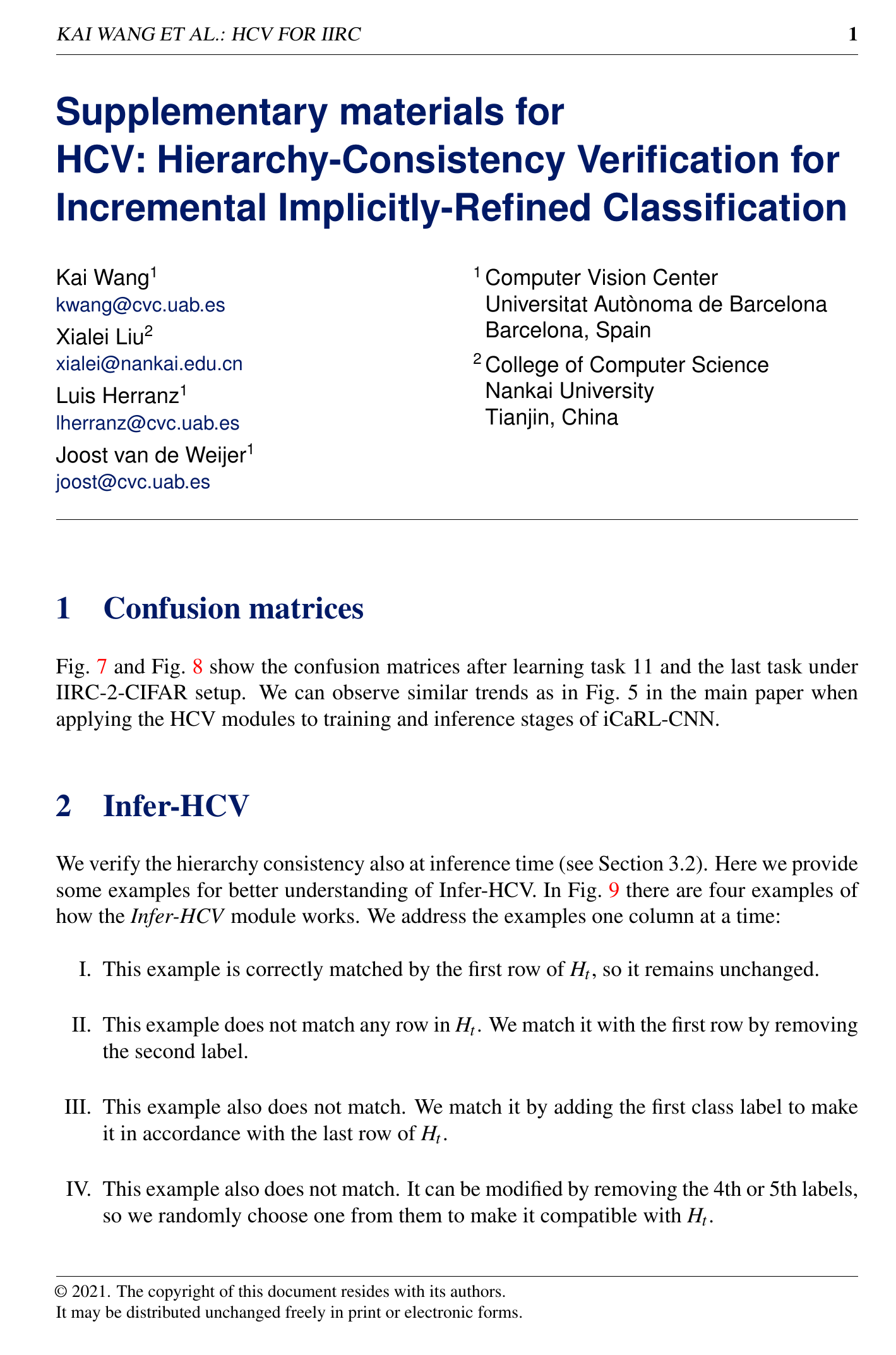}

\end{document}

\maketitle

%-------------------------------------------------------------------------
\setcounter{figure}{6}

\section{Confusion matrices}

\begin{figure*}[tb]
\begin{minipage}[b]{0.24\linewidth}
\centering
\includegraphics[width=\textwidth]{images/2_GT_confusion_matrix_10_icarl_cnn_ptm-crop.pdf}
\subcaption{Ground truth}
\label{fig:2_GT_confusion_matrix_10_icarl_cnn_ptm}
\end{minipage}
% \hspace{5mm}
\begin{minipage}[b]{0.24\linewidth}
\centering
\includegraphics[width=\textwidth]{images/2_ori_confusion_matrix_10_icarl_cnn_ptm-crop.pdf}
\subcaption{iCaRL-CNN}
\label{fig:2_ori_confusion_matrix_10_icarl_cnn_ptm}
\end{minipage}
% \hspace{5mm}
\begin{minipage}[b]{0.24\linewidth}
\centering
\includegraphics[width=\textwidth]{images/2_train_confusion_matrix_10_icarl_cnn_ptm-crop.pdf}
\subcaption{+ SPL}
\label{fig:2_train_confusion_matrix_10_icarl_cnn_ptm}
\end{minipage}
% \hspace{5mm}
\begin{minipage}[b]{0.24\linewidth}
\centering
\includegraphics[width=\textwidth]{images/2_both_confusion_matrix_10_icarl_cnn_ptm-crop.pdf}
\subcaption{+SPL+HCV}
\label{fig:2_both_confusion_matrix_10_icarl_cnn_ptm}
\end{minipage}

\begin{minipage}[b]{0.24\linewidth}
\centering
\includegraphics[width=\textwidth]{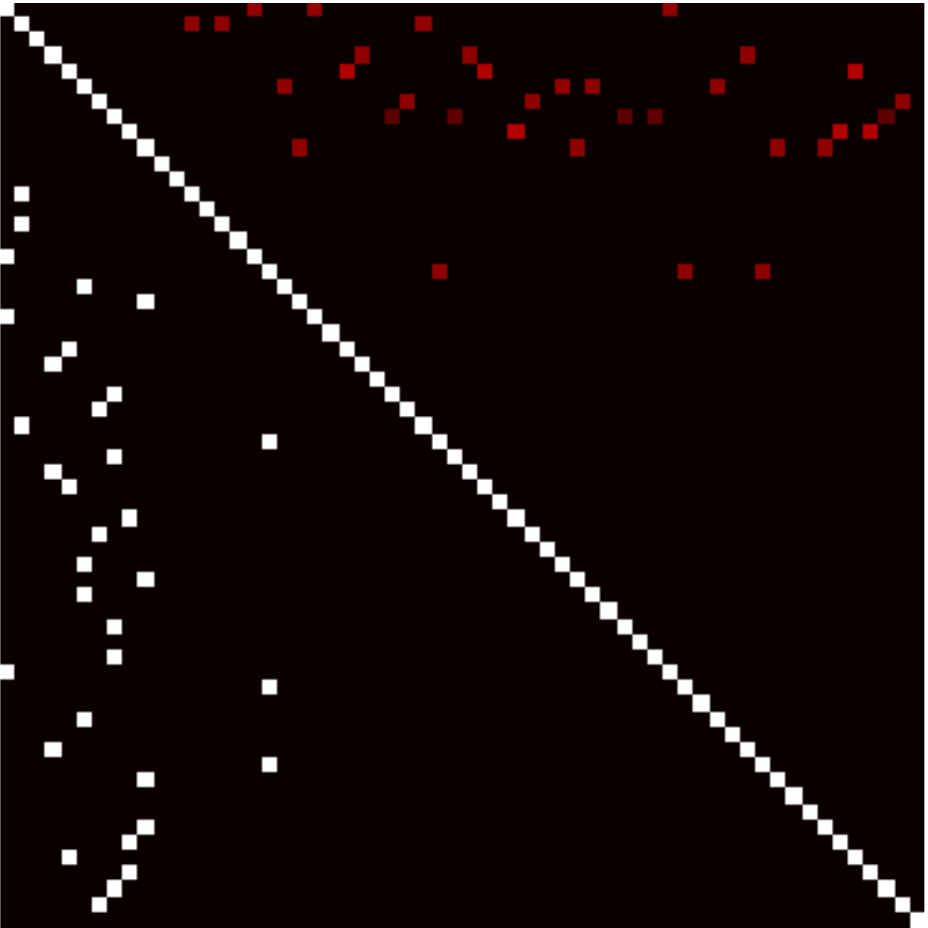}
\subcaption{Ground truth}
\label{fig:2_GT_confusion_matrix_10_lucir}
\end{minipage}
% \hspace{5mm}
\begin{minipage}[b]{0.24\linewidth}
\centering
\includegraphics[width=\textwidth]{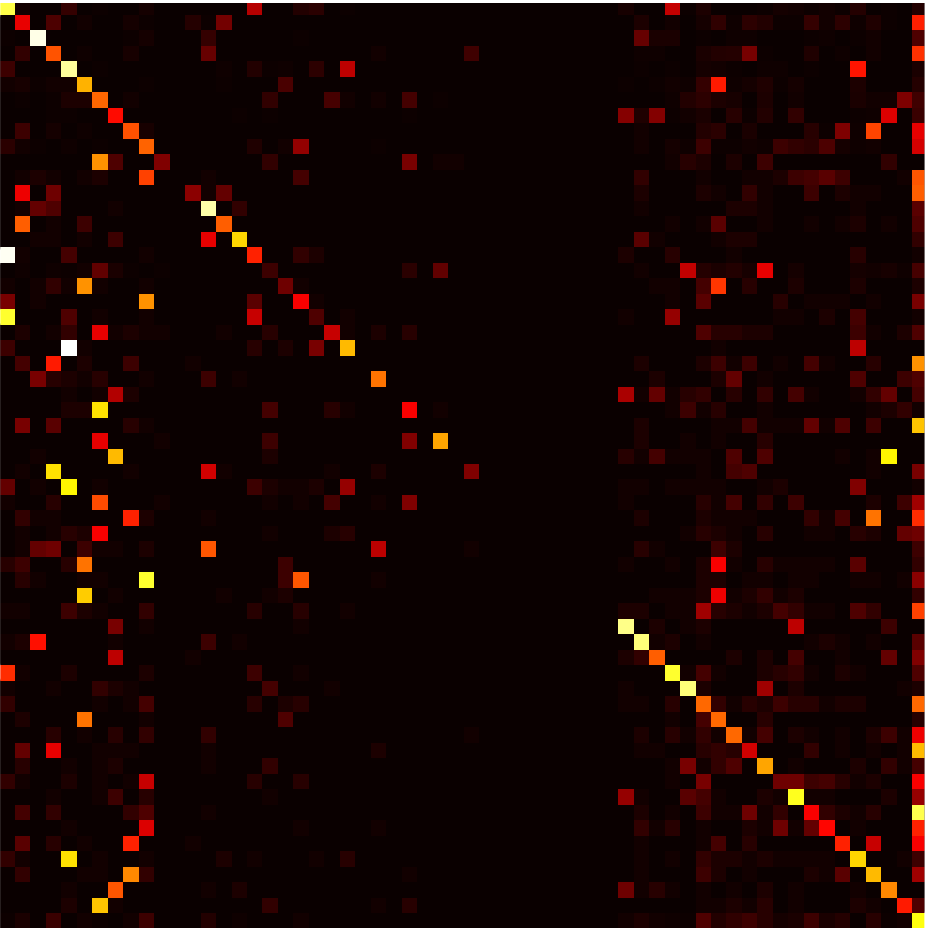}
\subcaption{LUCIR}
\label{fig:2_ori_confusion_matrix_10_lucir}
\end{minipage}
% \hspace{5mm}
\begin{minipage}[b]{0.24\linewidth}
\centering
\includegraphics[width=\textwidth]{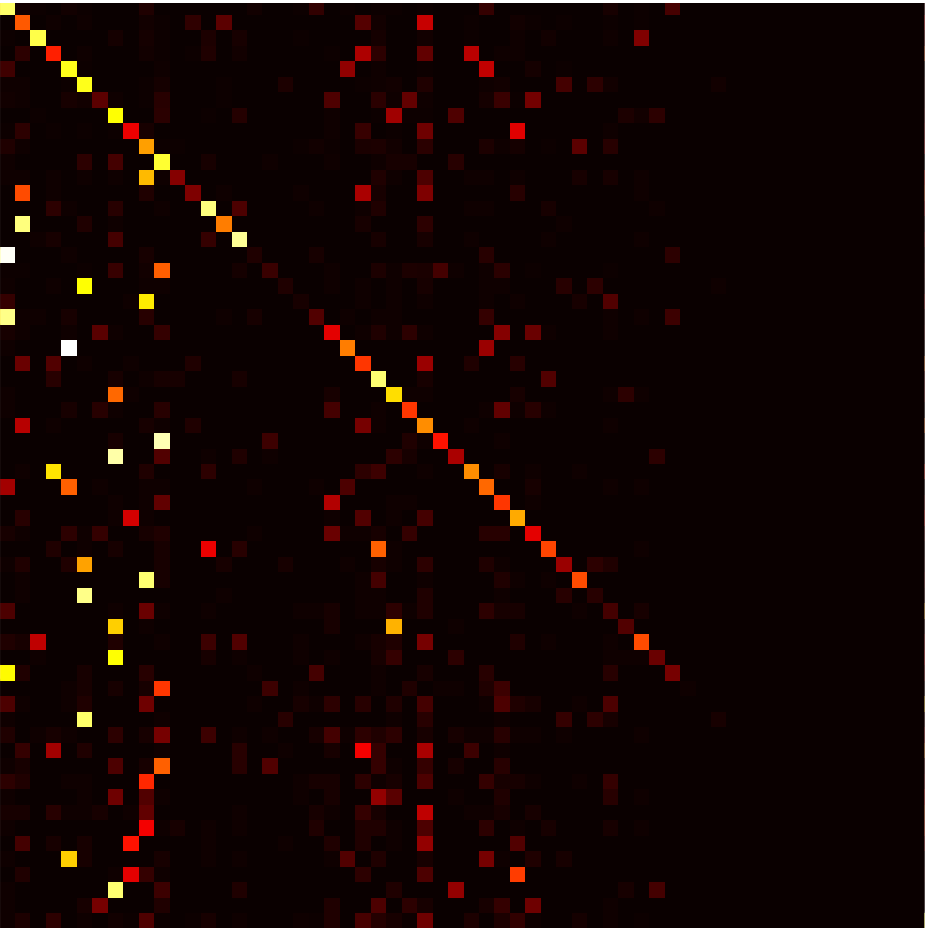}
\subcaption{+ SPL}
\label{fig:2_train_confusion_matrix_10_lucir}
\end{minipage}
% \hspace{5mm}
\begin{minipage}[b]{0.24\linewidth}
\centering
\includegraphics[width=\textwidth]{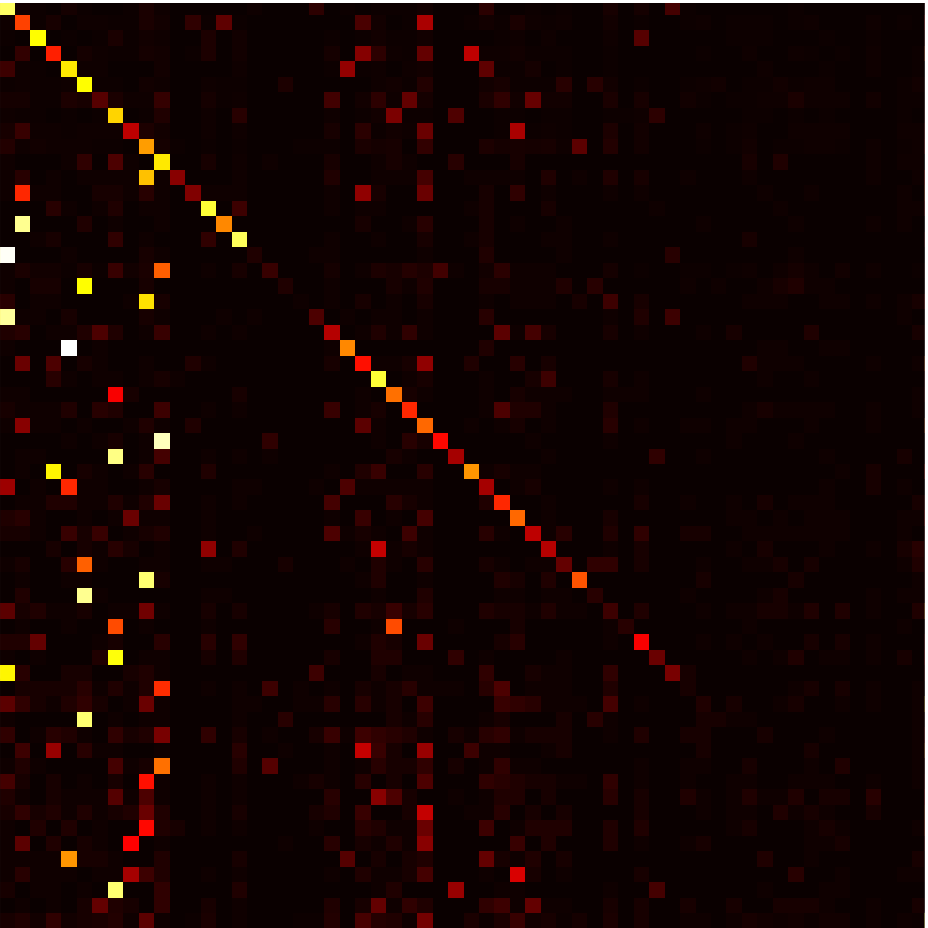}
\subcaption{+SPL+HCV}
\label{fig:2_both_confusion_matrix_10_lucir}
\end{minipage}

% \vspace{-2mm}
\caption{Confusion matrices of groundtruth, original continual learning methods, applying SPL and applying Infer-HCV after task 11 under IIRC-2-CIFAR setup. The first row is obtained with iCaRL-CNN as the base method and the second row is based on LUCIR.
% \vspace{-5mm}
}
% \vspace{-3mm}
\label{fig:confusion_matrices_11}
\end{figure*}

\begin{figure*}[htbp!]
\begin{minipage}[b]{0.24\linewidth}
\centering
\includegraphics[width=\textwidth]{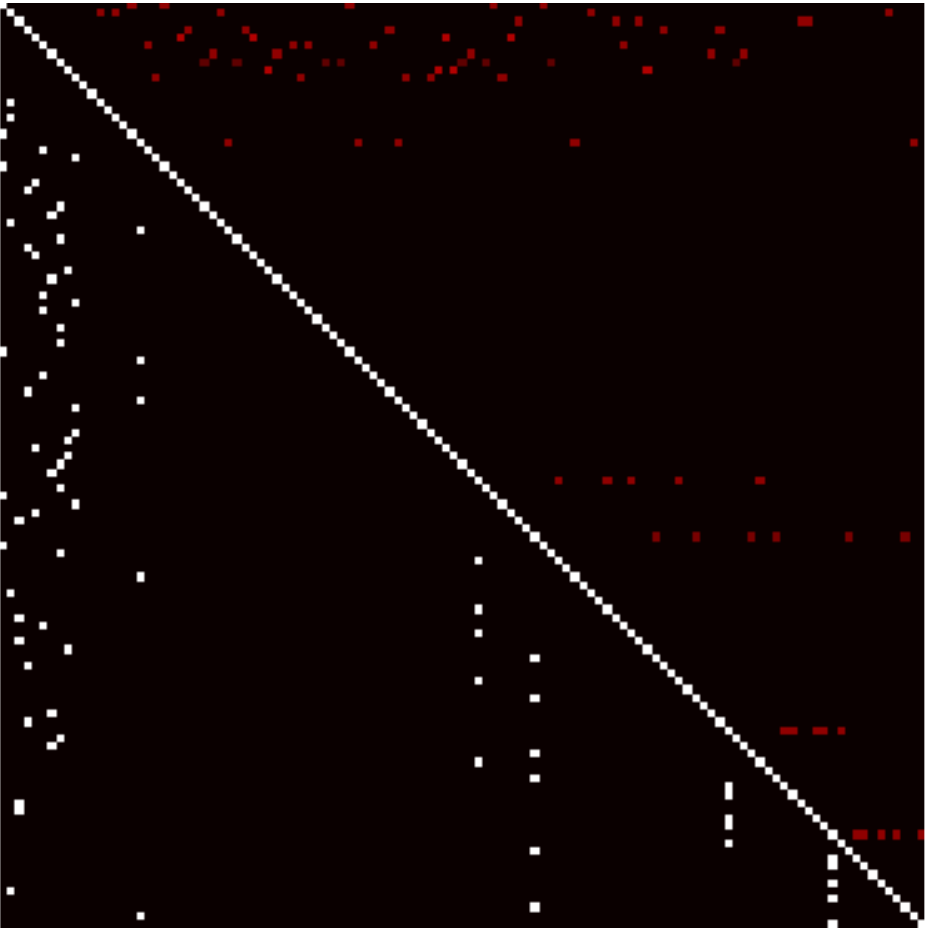}
\subcaption{Ground truth}
\label{fig:2_GT_confusion_matrix_21_icarl_cnn_ptm}
\end{minipage}
% \hspace{-21mm}
\begin{minipage}[b]{0.24\linewidth}
\centering
\includegraphics[width=\textwidth]{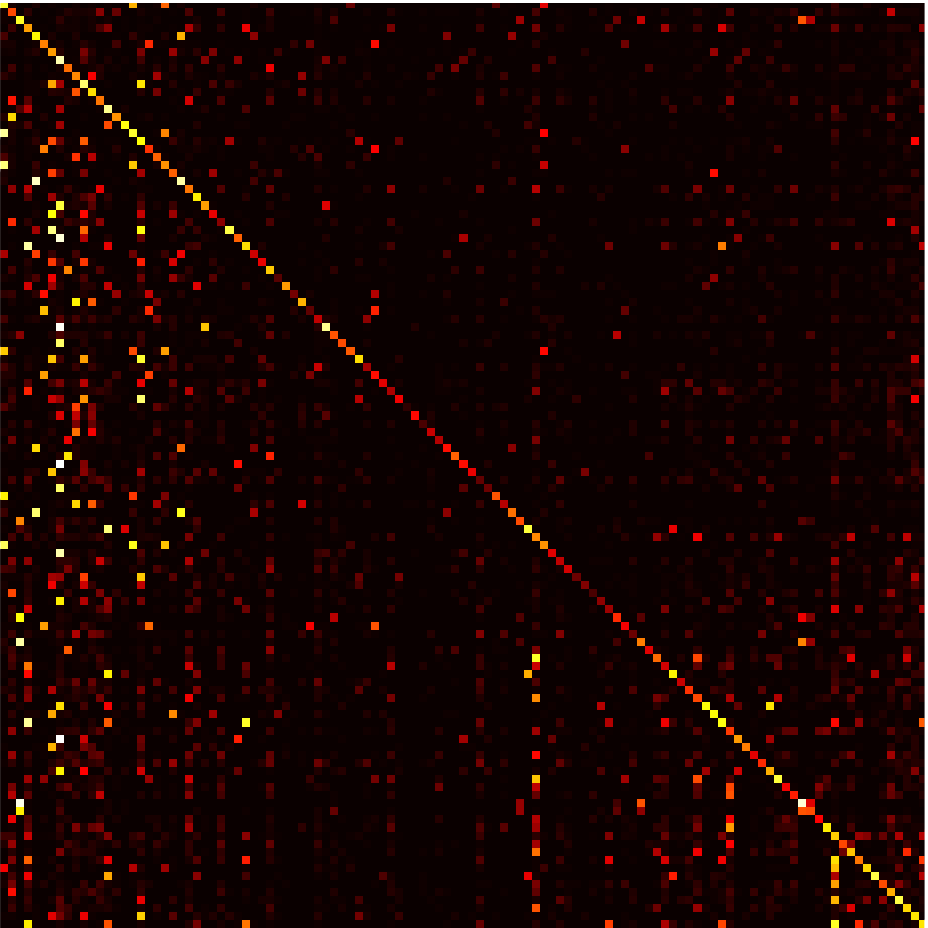}
\subcaption{iCaRL-CNN}
\label{fig:2_ori_confusion_matrix_21_icarl_cnn_ptm}
\end{minipage}
% \hspace{-21mm}
\begin{minipage}[b]{0.24\linewidth}
\centering
\includegraphics[width=\textwidth]{images/2_test_confusion_matrix_21_icarl_cnn_ptm-crop.pdf}
\subcaption{+ SPL}
\label{fig:2_train_confusion_matrix_21_icarl_cnn_ptm}
\end{minipage}
\begin{minipage}[b]{0.24\linewidth}
\centering
\includegraphics[width=\textwidth]{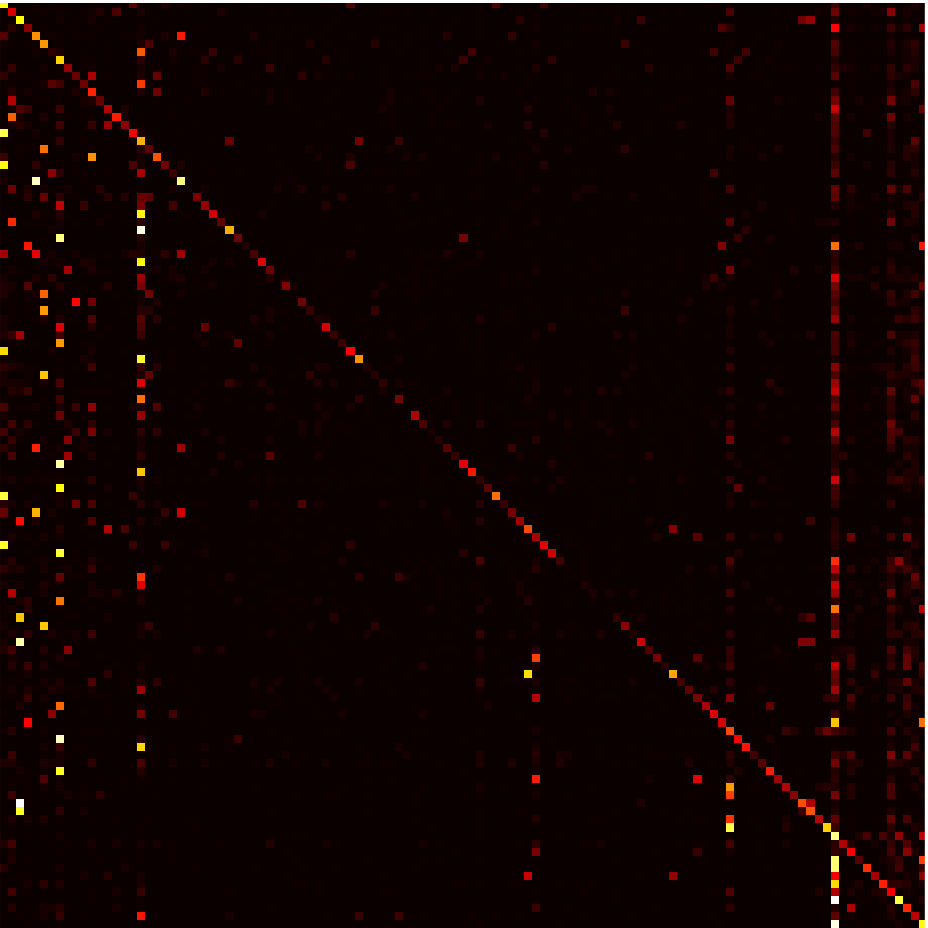}
\subcaption{+ SPL + Infer-HCV}
\label{fig:2_both_confusion_matrix_21_icarl_cnn_ptm}
\end{minipage}

\begin{minipage}[b]{0.24\linewidth}
\centering
\includegraphics[width=\textwidth]{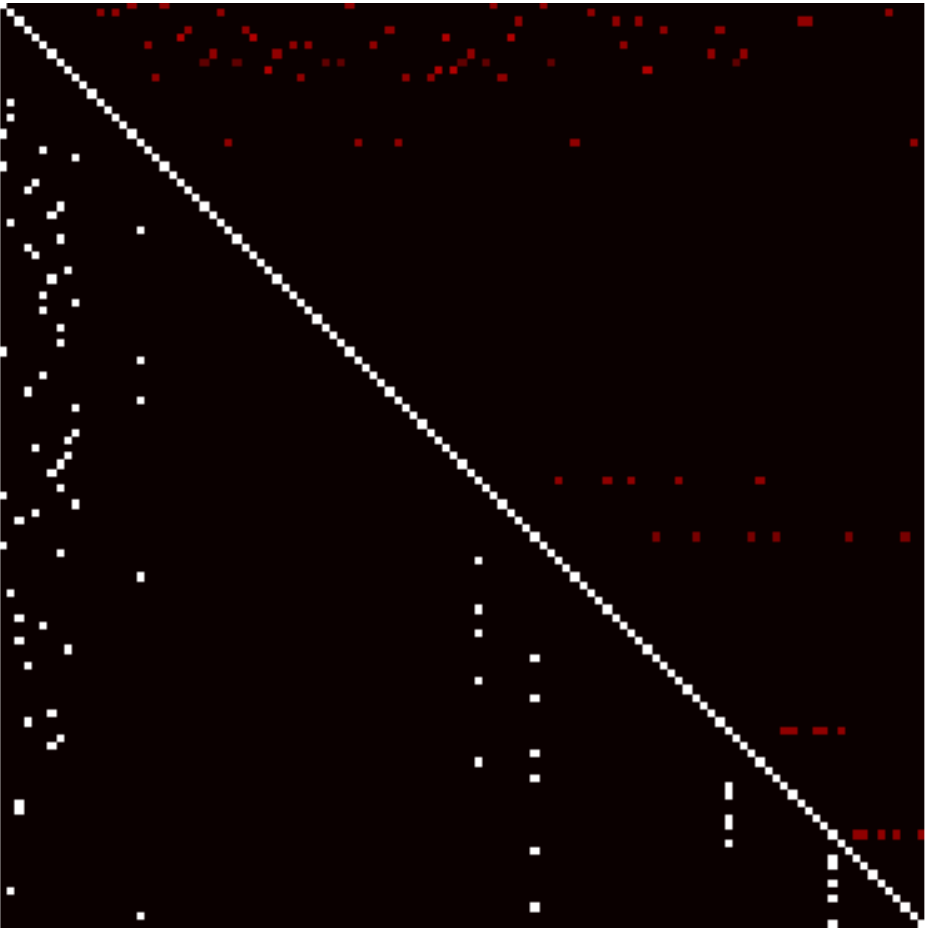}
\subcaption{Ground truth}
\label{fig:2_GT_confusion_matrix_21_lucir}
\end{minipage}
% \hspace{-21mm}
\begin{minipage}[b]{0.24\linewidth}
\centering
\includegraphics[width=\textwidth]{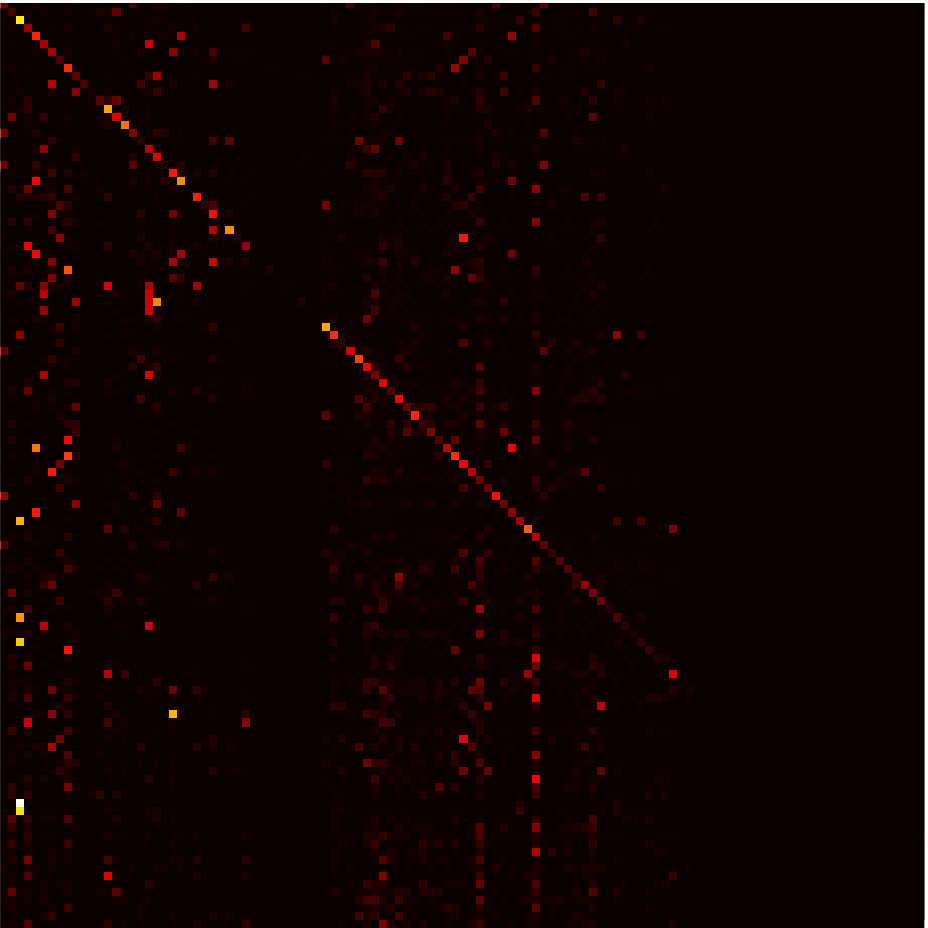}
\subcaption{LUCIR}
\label{fig:2_ori_confusion_matrix_21_lucir}
\end{minipage}
% \hspace{-21mm}
\begin{minipage}[b]{0.24\linewidth}
\centering
\includegraphics[width=\textwidth]{images/2_test_confusion_matrix_21_lucir-crop.pdf}
\subcaption{+ SPL}
\label{fig:2_train_confusion_matrix_21_lucir}
\end{minipage}
\begin{minipage}[b]{0.24\linewidth}
\centering
\includegraphics[width=\textwidth]{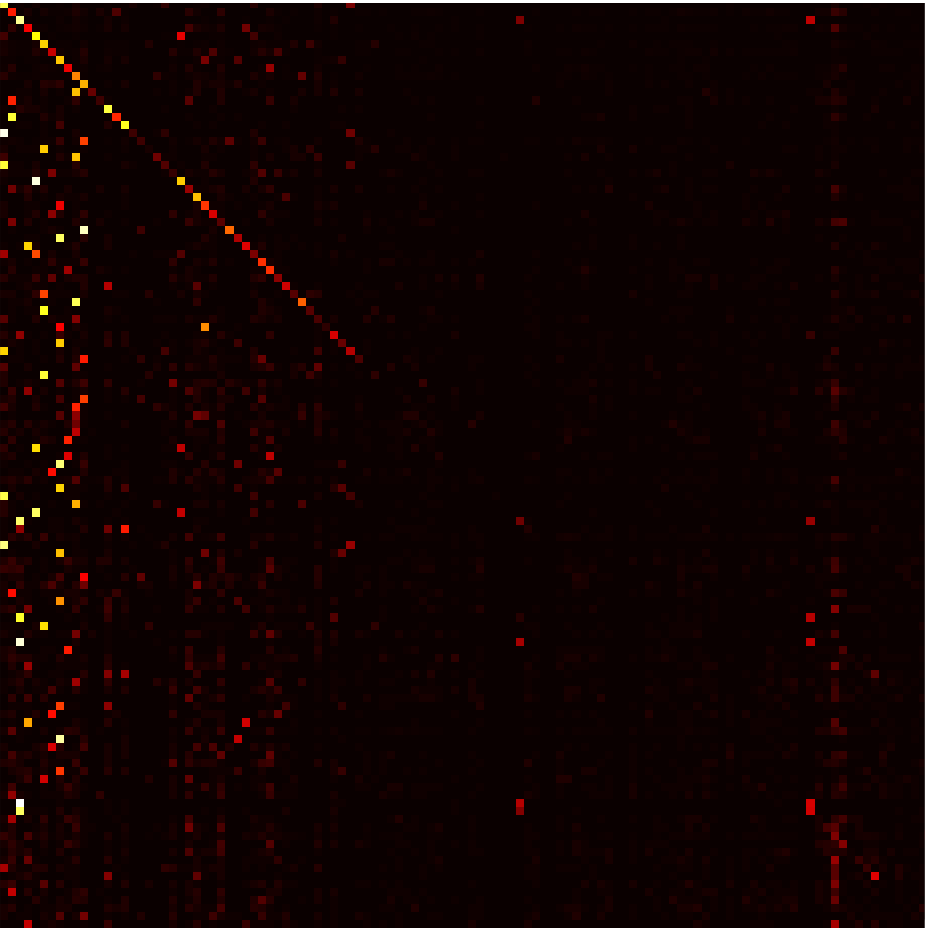}
\subcaption{+ SPL + Infer-HCV}
\label{fig:2_both_confusion_matrix_21_lucir}
\end{minipage}

% \vspace{-2mm}
\caption{Confusion matrices of groundtruth, original continual learning methods, applying SPL and applying Infer-HCV after the last task under IIRC-2-CIFAR setup. The first row is obtained with iCaRL-CNN as the base continual learning method and the second row is based on LUCIR.
% \vspace{-2mm}
}
% \vspace{-3mm}
\label{fig:confusion_matrices}
\end{figure*}

Fig.~\ref{fig:confusion_matrices_11} and Fig.~\ref{fig:confusion_matrices} show the confusion matrices after learning task 11 and the last task under IIRC-2-CIFAR setup. We can observe similar trends as in Fig.~5 in the main paper when applying the HCV modules to training and inference stages of iCaRL-CNN.

\section{Infer-HCV}

\begin{figure*}[htbp!]
\begin{center}
\includegraphics[width=0.99\textwidth]{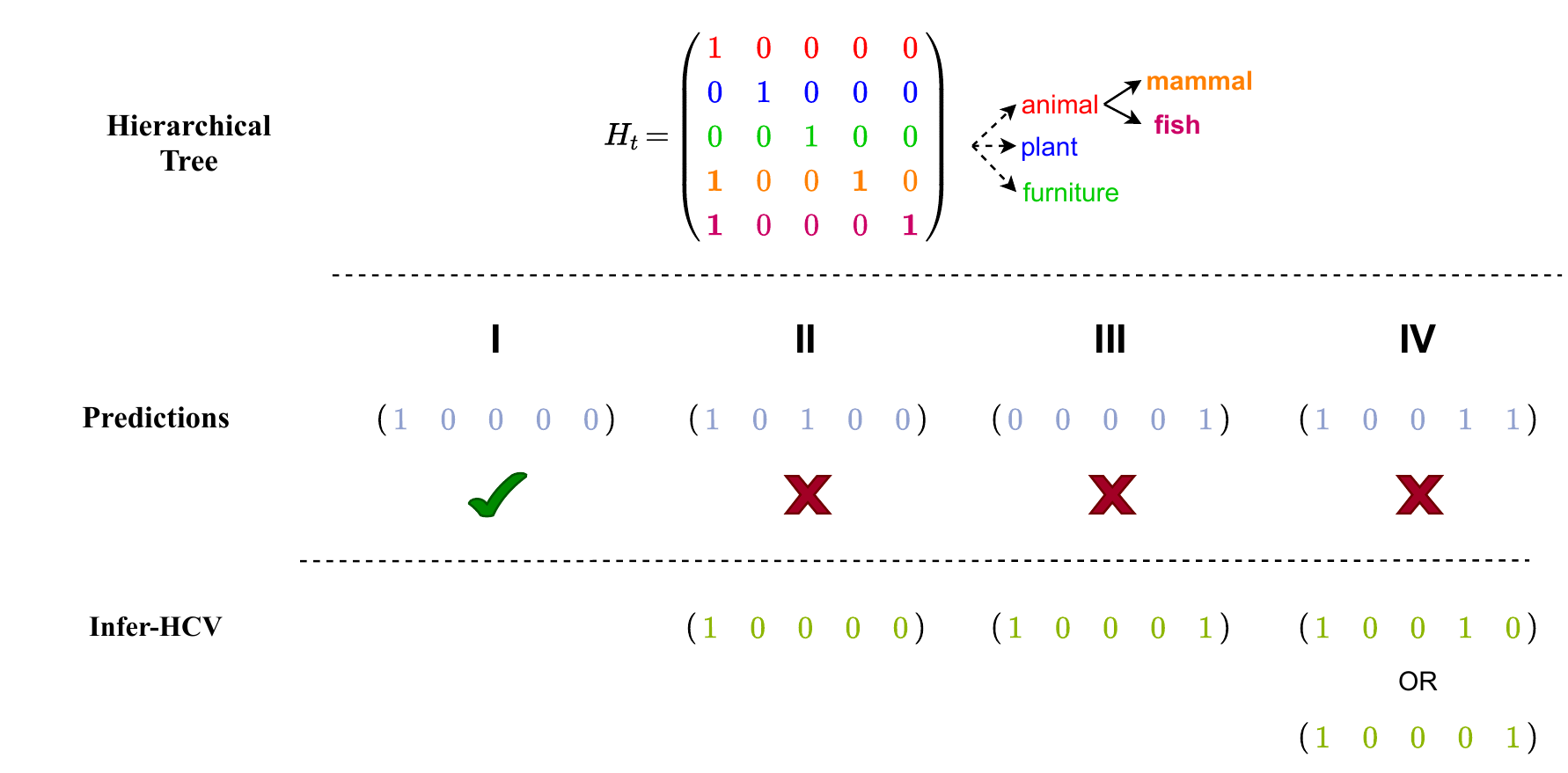}
\end{center}
\vspace{-2mm}
   \caption{Examples Illustration of Infer-HCV procedure. 
   \vspace{-2mm}
   }
  \vspace{-2mm}
\label{fig:infer_hcv}
\end{figure*}

We verify the hierarchy consistency also at inference time (see Section 3.2). Here we provide some examples for better understanding of Infer-HCV. In Fig.~\ref{fig:infer_hcv} there are four examples of how the \textit{Infer-HCV} module works. We address the examples one column at a time:
\begin{enumerate}[I.]
    \item This example is correctly matched by the first row of $H_t$, so it remains unchanged.
    \item This example does not match any row in $H_t$. We match it with the first row by removing the second label.
    \item This example also does not match. We match it by adding the first class label to make it in accordance with the last row of $H_t$.
    \item This example also does not match. It can be modified by removing the 4th or 5th labels, so we randomly choose one from them to make it compatible with $H_t$.
\end{enumerate}

\section{Groundtruth hierarchies and task splits}
Here we show the groundtruth hierarchies of IIRC-2-CIFAR/IIRC-3-CIFAR in Table~\ref{tab:iirc_3_cifar_hierarchy} and IIRC-ImageNet-Subset in Table~\ref{tab:iirc_imagenet_subset_hierarchy}. For the task splits used in our experiments, we select one from the IIRC paper for IIRC-2-CIFAR (Table~\ref{tab:iirc_2_task_configuration}) and propose our splits for IIRC-3-CIFAR (Tab.~\ref{tab:iirc_3_task_configuration})/IIRC-ImageNet-Subset (Tab.~\ref{tab:iirc_imagenet_subset_task_configuration}) to test the models in more complex hierarchies.

\section{Experimental results for IIRC-ImageNet-full}

\begin{figure*}[htbp!]
\begin{center}
\includegraphics[width=.75\textwidth]{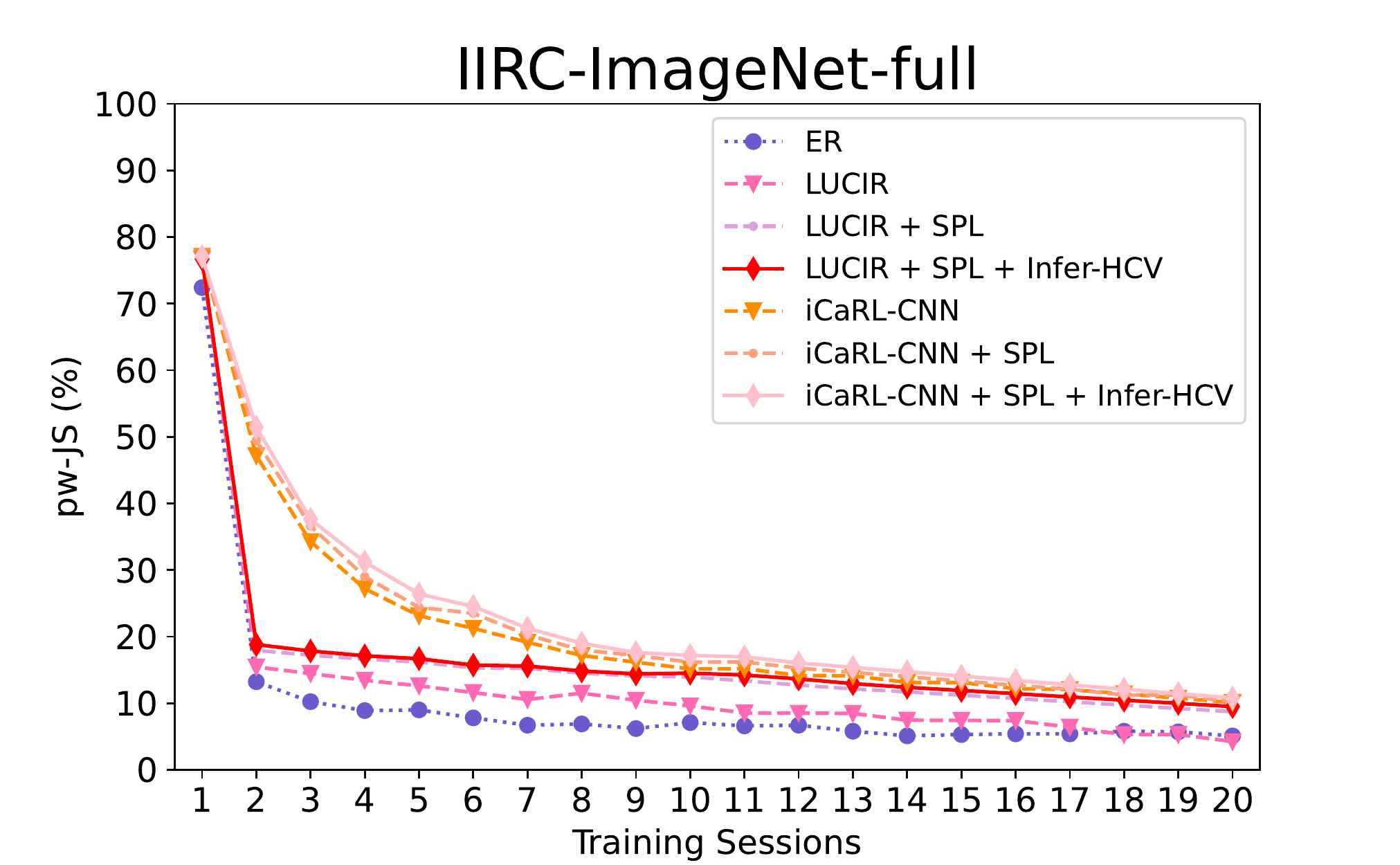}
\end{center}
\vspace{-3mm}
  \caption{Experiments for IIRC-ImageNet-full. 
  \vspace{-3mm}
  }
%   \vspace{-1mm}
\label{fig:rebuttal_iirc_imagenet_full_13task_LUCIR}
\end{figure*}

In Fig.~\ref{fig:rebuttal_iirc_imagenet_full_13task_LUCIR} we show results for the first 20 tasks (out of 35 in total)\footnote{Because of the demanding computational resources required for this experiment we were only able to compute results until task 20.} for LUCIR and iCaRL-CNN. Our HCV improves by 4.2\% for LUCIR and 1.9\% for iCaRL-CNN after 20 tasks.

\begin{table}[htbp!]
  \begin{center}
  \scalebox{0.89}
  {
  \begin{tabular}{|c|c|c|}
  \hline
  rootclass & superclass & subclasses \\
  \hline
  \hline
    \multirow{8}{*}{animals} & aquatic mammals & beaver,dolphin,otter,seal,whale \\
    \cline{2-3}
     & fish & aquarium fish,flatfish,ray,shark,trout \\
     \cline{2-3}
     & insects & bee,beetle,butterfly,caterpillar,cockroach \\
     \cline{2-3}
     & large carnivores & leopard,lion,tiger,wolf \\
     \cline{2-3}
     & large omnivores and herbivores & bear,camel,cattle,chimpanzee,elephant,kangaroo\\
     \cline{2-3}
     & medium sized mammals & fox,porcupine,possum,raccoon,skunk\\
     \cline{2-3}
     & reptiles & crocodile,dinosaur,lizard,snake,turtle\\
     \cline{2-3}
     & small mammals & hamster,mouse,rabbit,shrew,squirrel\\
    %  \cline{2-3}
     
    \hline
    \multirow{3}{*}{plants} & flowers & orchid,poppy,rose,sunflower,tulip\\
     \cline{2-3}
     & fruit and vegetables & apple,orange,pear,sweet pepper\\
     \cline{2-3}
     & trees & maple tree,oak tree,palm tree,pine tree,willow tree\\
    %  \cline{2-3}
    \hline
    
     \multirow{5}{*}{ - }& \multirow{2}{*}{vehicles} & bicycle,bus,motorcycle,pickup truck, \\ 
     & & train,streetcar,tank,tractor\\
     \cline{2-3}
     & household furniture & bed,chair,couch,table,wardrobe\\
     \cline{2-3}
     & food containers & bottle,bowl,can,cup,plate\\
     \cline{2-3}
     & people & baby,boy,girl,man,woman\\
    %  \cline{2-3}
    \hline
    
    \multirow{4}{*}{ - }& \multirow{4}{*}{ - } & mushroom,clock,keyboard,lamp,telephone, \\ 
     &&television,bridge,castle,house,road, \\ 
     &&skyscraper,cloud,forest,mountain,plain, \\  &&sea,crab,lobster,snail,spider,worm,lawn mower,rocket \\
     \hline
     
  \end{tabular}
  }
  \end{center}
  \caption{IIRC-2-CIFAR and IIRC-3-CIFAR Hierarchies. IIRC-2-CIFAR is without the rootclasses.}
  \label{tab:iirc_3_cifar_hierarchy}
\end{table}

\begin{table}[htbp!]
  \begin{center}
  \begin{tabular}{|c|c|}
  \hline
  superclass & subclasses \\
  \hline
  \hline
    \multirow{3}{*}{primate} & Madagascar cat, indri, gibbon, siamang, orangutan, gorilla, chimpanzee,  \\
    & marmoset, capuchin, howler monkey,titi, spider monkey, squirrel monkey, \\
    & guenon, patas, baboon, macaque, langur, colobus, proboscis monkey  \\

    \hline
    
    \multirow{2}{*}{car} & ambulance, beach wagon, cab, convertible, jeep, limousine, \\
    & Model T, racer, sports car, minivan, grille, golfcart  \\
     \hline
    \multirow{2}{*}{ball} & baseball, basketball, croquet ball, golf ball, ping-pong ball,\\
    &  punching bag, rugby ball, soccer ball, tennis ball, volleyball  \\
    \hline
    \multirow{2}{*}{crustacean} & isopod, crayfish, hermit crab, spiny lobster, American lobster, \\
    & Dungeness crab, rock crab, fiddler crab, king crab  \\
    \hline
    \multirow{2}{*}{weapon} & bow, projectile, cannon, missile, rifle, \\
    & revolver, assault rifle, holster bison, \\
    \hline
    \multirow{2}{*}{spider} & black and gold garden spider, barn spider, garden spider,\\
    & black widow, tarantula, wolf spider, spider web \\
    \hline
    box & carton, chest, crate, mailbox, pencil box, safe \\
    \hline
    bag & backpack, mailbag, plastic bag, purse, sleeping bag \\
    \hline
    beverage & espresso, red wine, cup, eggnog \\
    \hline
    keyboard & computer keyboard, space bar, typewriter keyboard \\
    \hline
     \multirow{3}{*}{-} & African crocodile, American alligator, triceratops, trilobite, harvestman,\\
     & scorpion, tick, centipede, tusker, echidna, platypus, jellyfish, \\
     & sea anemone, brain coral, flatworm, nematode \\
     \hline
    
  \end{tabular}
  \end{center}
  \caption{IIRC-ImageNet-Subset Hierarchy}
  \label{tab:iirc_imagenet_subset_hierarchy}
\end{table}

\begin{table}[htbp!]
\begin{center}
\begin{tabular}{|c|c|}
\hline
Task ID & Classes \\
\hline\hline
\multirow{3}{*}{1} & flowers \textbf{(S)}, small mammals \textbf{(S)}, trees \textbf{(S)},  \\& aquatic mammals \textbf{(S)}, fruit and vegetables \textbf{(S)},  people \textbf{(S)}, \\ 
& food containers \textbf{(S)}, vehicles \textbf{(S)}, large carnivores \textbf{(S)}, insects \textbf{(S)} \\
\hline
2 & television, spider, shrew, mountain, hamster \\
\hline
3 & road, poppy, household furniture \textbf{(S)}, woman, bee \\
\hline
4 & tulip, clock, orange, beaver, rocket\\
\hline
5 & bicycle, can, squirrel, wardrobe, bus\\
\hline
6 & whale, sweet pepper, telephone, leopard, bowl\\
\hline
7 &skyscraper, baby, cockroach, boy, lobster\\
\hline
8&  motorcycle, forest, tank, orchid, chair\\
\hline
9 & crab, girl, keyboard, otter, bed\\
\hline
10 & butterfly, lawn mower, snail, caterpillar, wolf\\
\hline
11 & pear, tiger, pickup truck, cup, reptiles \textbf{(S)}\\
\hline
12 & train, sunflower, beetle, apple, palm tree\\
\hline
13 & plain, large omnivores and herbivores \textbf{(S)}, rose, tractor, crocodile\\
\hline
14 & mushroom, couch, lamp, mouse, bridge\\
\hline
15 & turtle, willow tree, man, lizard, maple tree\\
\hline
16 & lion, elephant, seal, sea, dinosaur\\
\hline
17 & worm, bear, castle, plate, dolphin\\
\hline
18 & medium sized mammals \textbf{(S)}, streetcar, bottle, kangaroo, snake\\
\hline
19 & house, chimpanzee, raccoon, porcupine, oak tree\\
\hline
20 & pine tree, possum, skunk, fish \textbf{(S)}, fox\\
\hline
21 & cattle, ray, aquarium fish, cloud, flatfish\\
\hline
22 & rabbit, trout, camel, table, shark\\

\hline
\end{tabular}
\end{center}
\caption{IIRC-2-CIFAR task split configuration. \textbf{(S)} denotes the super classes.}
\label{tab:iirc_2_task_configuration}
\end{table}

\begin{table}[htbp!]
\begin{center}
\begin{tabular}{|c|c|}
\hline
Task ID & Classes \\
\hline\hline
1 & \textbf{animals (R)}, food containers \textbf{(S)}, house, road, telephone, lamp, rocket \\
\hline
2 & fish \textbf{(S)}, crab, snail, mushroom, bottle \\
\hline
3 & reptiles \textbf{(S)}, sea, trout, can, clock \\
\hline
4 & vehicles \textbf{(S)}, flatfish, snake, aquarium fish, mountain\\
\hline
5 & large omnivores and herbivores \textbf{(S)}, plate, cup, television, ray\\
\hline
6 & small mammals \textbf{(S)}, camel, bicycle, keyboard, lawn mower\\
\hline
7 & household furniture \textbf{(S)}, shark, turtle, spider, forest\\
\hline
8&  aquatic mammals \textbf{(S)}, couch, kangaroo, streetcar, lobster\\
\hline
9 & insects \textbf{(S)}, tank, plain, motorcycle, whale\\
\hline
10 & large carnivores \textbf{(S)}, bear, crocodile, dinosaur, otter\\
\hline
11 & \textbf{plants (R)}, medium sized mammals \textbf{(S)}, tractor, table, bowl\\
\hline
12 & fruit and vegetables \textbf{(S)}, chimpanzee, cloud, raccoon, butterfly\\
\hline
13 & flowers \textbf{(S)}, bus, shrew, wardrobe, apple\\
\hline
14 & people \textbf{(S)}, mouse, skunk, caterpillar, castle\\
\hline
15 & trees \textbf{(S)}, fox, cockroach, worm, squirrel\\
\hline
16 & palm tree, oak tree, lizard, maple tree, orchid\\
\hline
17 & rose, bed, train, elephant, sweet pepper\\
\hline
18 & skyscraper, beetle, boy, pine tree, lion\\
\hline
19 & hamster, pear, bee, man, tiger\\
\hline
20 & seal, poppy, porcupine, dolphin, tulip\\
\hline
21 & orange, wolf, cattle, willow tree, woman\\
\hline
22 & bridge, pickup truck, baby, sunflower, girl\\
\hline
23 & leopard, rabbit, beaver, possum, chair\\

\hline
\end{tabular}
\end{center}
\caption{IIRC-3-CIFAR task split configuration. \textbf{(R)} denotes the root classes and \textbf{(S)} denotes the super classes.}
\label{tab:iirc_3_task_configuration}
\end{table}

\begin{table}[htbp!]
\begin{center}
\begin{tabular}{|c|c|}
\hline
Task ID & Classes \\
\hline\hline
\multirow{2}{*}{1} & car \textbf{(S)}, box \textbf{(S)}, brain coral, echidna, American alligator,  \\& African crocodile, tusker, jellyfish, sea anemone, tick\\
\hline
\multirow{2}{*}{2}  &bag \textbf{(S)}, crustacean \textbf{(S)}, trilobite, ambulance, triceratops, \\& sports car, mailbox, grille, nematode, limousine\\
\hline
\multirow{2}{*}{3} & weapon \textbf{(S)}, pencil box, jeep, Model T, purse, \\& mailbag, hermit crab, centipede, beach wagon, fiddler crab\\
\hline
\multirow{2}{*}{4} & beverage \textbf{(S)}, safe, minivan, scorpion, assault rifle, \\& harvestman, missile, sleeping bag, holster, projectile\\
\hline
\multirow{2}{*}{5} & keyboard \textbf{(S)}, revolver, Dungeness crab, American lobster, king crab, \\& cab, convertible, racer, isopod, chest\\
\hline
\multirow{2}{*}{6} & ball \textbf{(S)}, backpack, cannon, crayfish, rock crab, \\& espresso, platypus, flatworm, rifle, space bar\\
\hline
\multirow{2}{*}{7} & spider \textbf{(S)}, spiny lobster, volleyball, punching bag, carton, \\& golfcart, plastic bag, golf ball, eggnog, baseball\\
\hline
\multirow{2}{*}{8} & primate \textbf{(S)}, black widow, basketball, typewriter keyboard, bow, \\& tarantula, garden spider, rugby ball, cup, black and gold garden spider\\
\hline
\multirow{2}{*}{9} & wolf spider, marmoset, squirrel,  monkey, guenon, \\& orangutan, macaque, baboon, Madagascar cat, capuchin, soccer ball\\
\hline
\multirow{2}{*}{10} & howler monkey, siamang, gibbon, gorilla, spider web, \\& red wine, crate, colobus, tennis ball, barn spider\\
\hline
\multirow{2}{*}{11} & croquet ball, indri, chimpanzee, titi, spider monkey, \\& langur, ping-pong ball, computer keyboard, patas, proboscis monkey\\
\hline

\hline
\end{tabular}
\end{center}
\caption{IIRC-ImageNet-Subset task split configuration. \textbf{(S)} denotes the super classes.}
\label{tab:iirc_imagenet_subset_task_configuration}
\end{table}